\pdfoutput=1
\documentclass[11pt]{article}

\usepackage{acl}

\usepackage{times}
\usepackage{latexsym}

\usepackage[T1]{fontenc}
\usepackage[utf8]{inputenc}

\usepackage{microtype}

\usepackage{sec/package}

\usepackage{color}
\usepackage{soul}
\usepackage{todonotes}

\definecolor{pigment}{rgb}{0.2, 0.2, 0.6}

\definecolor{blue}{RGB}{0, 93, 170}			%Go Big Blue!
\definecolor{darkgreen}{HTML}{3bb35b}
\newcommand{\logicGeneral}{\textsc{{Logic}}}
\newcommand{\logicClimate}{\textsc{{LogicClimate}}}

\newcommand{\electra}{{Electra}}
\newcommand{\ourmodel}{Electra-\textit{StructAware}}

\definecolor{darkyellow}{HTML}{f0b71d}

\newcommand{\ignore}[1]{}
\newcommand*\samethanks[1][\value{footnote}]{\footnotemark[#1]}
 \usepackage{multirow}
 \usepackage{float}
 \usepackage{physics}

 \usepackage{bbm}
 
 \usepackage{amsmath}
\title{
Logical Fallacy Detection
}

\author{
Zhijing Jin\textsuperscript{\rm 1,2,}\thanks{\hspace{0.1cm} Equal contribution.} \hspace{0.2cm}
Abhinav Lalwani\textsuperscript{\rm 3,}\samethanks \hspace{0.03cm} \textsuperscript{\rm ,}\thanks{\hspace{0.1cm} Done during the research internship at ETH Zürich.} \hspace{0.2cm}
Tejas Vaidhya\textsuperscript{\rm 4,}\samethanks \hspace{0.3cm}
Xiaoyu Shen\textsuperscript{\rm 5} \hspace{0.1cm}
Yiwen Ding\textsuperscript{\rm 6} \hspace{0.2cm}
\\  % All authors must be in the same font size and format. Use \Large and \textbf to achieve this result when breaking a line
{\bf Zhiheng Lyu\textsuperscript{\rm 7,}\samethanks \hspace{0.3cm}
Mrinmaya Sachan\textsuperscript{\rm 2} \hspace{0.2cm}
Rada Mihalcea\textsuperscript{\rm 6} \and Bernhard Schölkopf\textsuperscript{\rm 1,2}
}
\\
\textsuperscript{\rm 1}Max Planck Institute,
\textsuperscript{\rm 2}ETH Zürich, \textsuperscript{\rm 3}BITS Pilani, 
\textsuperscript{\rm 4}IIT Kharagpur,
\\
\textsuperscript{\rm 5}Saarland Informatics Campus, 
\textsuperscript{\rm 6}University of Michigan,
\textsuperscript{\rm 7}University of Hong Kong \\
\texttt{jinzhi@ethz.ch} \quad \texttt{abhinav.lalwani@gmail.com}
}
\begin{document}
\maketitle

\begin{abstract}
Reasoning is central to human intelligence. However, fallacious arguments are common, and some exacerbate problems such as spreading misinformation about climate change. In this paper, we propose the task of \textit{logical fallacy detection}, and provide a new dataset (\textbf{\logicGeneral{}}) of logical fallacies generally found in text, together with an additional challenge set for detecting logical fallacies in climate change claims (\textbf{\logicClimate{}}). Detecting logical fallacies is a hard problem as the model must understand the underlying logical structure of the argument. We find that existing pretrained large language models perform poorly on this task. 
In contrast, we show that a simple structure-aware classifier outperforms the best language model by 5.46\% $F_1$ scores on \logicGeneral{} and 4.51\% on \logicClimate{}. We encourage future work to explore this task since (a) it can serve as a new reasoning challenge for language models, and (b) it can have potential applications in tackling the spread of misinformation.\footnote{
Our dataset and code 
% will be available upon acceptance.
are available at \url{https://github.com/causalNLP/logical-fallacy}.
}
\end{abstract}

\section{Introduction}
Reasoning is the process of using existing knowledge to make inferences, create explanations, and generally assess things rationally by using logic \cite{aristotle1991rhetoric}. Human reasoning is, however, often marred with logical fallacies. Fallacious reasoning leads to disagreements, conflicts, endless debates, and a lack of consensus. In daily life, fallacious arguments can be as harmless as ``\textit{All tall people like cheese}'' (faulty generalization) or ``\textit{She is the best because she is better than anyone else}'' (circular claim). However, logical fallacies are also intentionally used to spread misinformation, for instance ``\textit{Today is so cold, so I don't believe in global warming}''
% the winter this year is colder, so we do not have any global warming
(faulty generalization) or ``\textit{Global warming doesn't exist because the earth is not getting warmer}'' (circular claim).
% \textcolor{red}{cite some claim verification papers}\textcolor{blue}{M, I cannot find claim verification papers without fact checking as an element.}

%The year of 2021 has seen a lot of abnormal weather changes, but climate change denial is still prevalent. To reach a consensus that can benefit society and 
In order to detect such fallacious arguments, we propose the task of \textit{logical fallacy detection}.
%to uncover the truth underlying the myriad different arguments.
%In this paper, we propose the new task of logical fallacy detection. 
Logical fallacy detection methods can be helpful to tackle important social problems. For instance, these methods can be combined with fact-checkers \cite{riedel2017simple,thorne-etal-2018-fever} %into a more extensive pipeline 
for misinformation detection as many claims can be factually correct but still fallacious.
However, logical fallacy detection is challenging as it requires a model to discover egregious patterns of reasoning \cite{johnson2006logical,damer2009attacking}.

\begin{figure}[t]
\centering
    \includegraphics[width=\columnwidth]{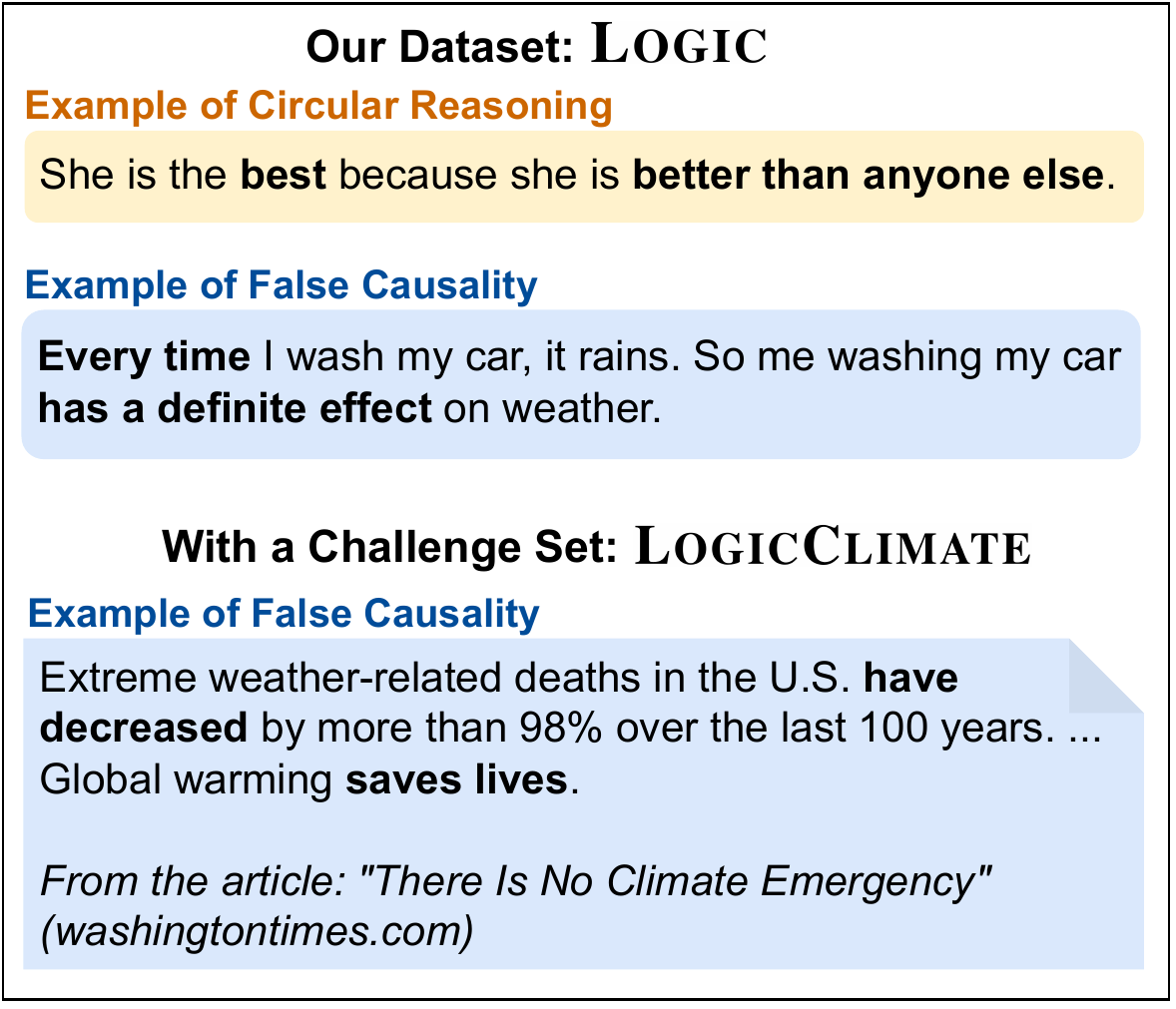}
    \caption{Our dataset consists of general logical fallacies (\logicGeneral{}) and an additional test set of logical fallacies in climate claims (\logicClimate{}).} \label{fig:intro}
\end{figure}
\begin{table*}[t]
    \centering
    \small
    \begin{tabular}{p{2.2cm}p{13.2cm}
    % p{5cm}lllllll
    }
\toprule
\textbf{Logical Fallacy} & \textbf{Examples}
% & \textbf{Common Pattern} 
\\ \hline
\textbf{Faulty Generalization} (18.01\%)
& ``I met \textbf{a tall man} who loved to eat cheese. Now I believe that \textbf{all tall people} like cheese.'' 
% & Set-A has a Property-X because its Proper-Subset-B has the Property-X. 
\newline
\darkred{``\textbf{Sometimes} flu vaccines don't work; therefore \textbf{vaccines are useless}.''}
\\
% (18.01\%) & ``\textbf{An environmental group} illegally blocked loggers and workers at a nuclear plant. Therefore, \textbf{environmentalists are radicals} who take the law into their own hands.''
% \\ 
\hline
\textbf{Ad Hominem} 
& ``What can our new math teacher know? Have you seen \textbf{how fat} she is?'' 
% & Person-A's argument is bad because Person-A looks bad in an irrelevant 
\\

(12.33\%) 
% & ``Leonardo DiCaprio is just a \textbf{dumb actor}! {What does he really know} about climate change?!'' 
% \\
& \darkred{``I cannot listen to anyone \textbf{who does not share my social and political values}.''}
% Louise is running for class president. In her campaign speech she says, ``My opponent does not deserve to win. She is a smoker and she cheated on her boyfriend last year.'' 
\\ \hline
\textbf{Ad Populum} 
& ``Everyone should like coffee: \textbf{95\%} of teachers do!'' 
% & Either Subset-A or Subset-B, ignoring
\\ 
% McDonald's Hamburgers: over 99 billion served.
% Don't be the only one not wearing Nike!
(9.47\%) & \darkred{``Killing thousands of people as a result of drug war campaign is \textbf{not a crime} to humanity because \textbf{millions of Filipino support it}.''} 
% & other possibilities.
\\
% & ``In the court trial deliberations, Juror Seven says, `There are still \textbf{eleven of us} who think he's guilty. You're \textbf{alone}.'”
% \\
% & ``You \textbf{either} support Hillary Clinton for President \textbf{or} you don't believe in women's rights.'' \\
\hline
\textbf{False Causality}
& ``\textbf{Every time} I wash my car, it rains. Me washing my car has \textbf{a definite effect} on the weather.'' 
% & Event-A co-occurs with Event-B, so Event-A causes Event-B.
\\

(8.82\%) & \darkred{``Every severe recession \textbf{follows} a Republican Presidency; therefore Republicans \textbf{are the cause of} recessions.''} \\
% Before women got the vote, there were no nuclear weapons
% & ``The number of pirates in existence \textbf{decreased as the global average temperature rose}. Therefore, we can conclude that \textbf{hotter weather reduces piracy}.''
% ``The flu vaccine comes out every year \textbf{just before} the holidays, then people go and spend a lot of money. Therefore, businesses \textbf{are secretly using} the flu vaccine \textbf{as a mind control device} to increase holiday spending.''
% \\ 
\hline
\textbf{Circular Claim}
& ``J.K. Rowling is a \textbf{wonderful} writer because she \textbf{writes so well}.'' 
% & Claim-A is because of an equivalent Claim-B.
\\ 
(6.98\%) 
% & ``\textbf{Global warming doesn't exist} because the earth \textbf{is not getting warmer}.''
% \\ 
& \darkred{``She is the \textbf{best} candidate for president because she \textbf{is better than} the other candidates!''}
\\ \hline

\textbf{Appeal to \newline Emotion} 
& ``It is an \textbf{outrage} that the school wants to remove the vending machines. This is \textbf{taking our freedom away}!''
\\
(6.82\%) & \darkred{``Vaccines are so \textbf{unnatural}; it’s \textbf{disgusting} that people are willing to put something like that in their body.''}
\\ 
% & ``If you have never been born again, \textbf{eternal separation} from God in the \textbf{Lake of Fire awaits you}. If you are born again, then being with the Lord in heaven \textbf{forever is your destiny}. Which do you choose?'' \\
\hline

\textbf{Fallacy of} & ``\textbf{Why are you worried} about poverty? \textbf{Look how many} children we abort every day.''
\\
\textbf{Relevance}
\newline
(6.61\%) % & ``There is a lot of commotion regarding \textbf{saving the environment}. We \textbf{cannot make this world an Eden}. What will happen if it does become Eden? Adam and Eve got bored there!''
% \\
& \darkred{``\textbf{Why should we be worrying} about how the government treats Native people, \textbf{when} people in our city can't get a job''}
\\ \hline
\textbf{Deductive \newline Fallacy} & ``\textbf{It is possible} to fake the moon landing through special effects. \textbf{Therefore}, the moon landing was a fake using special effects.''\\ 
(6.21\%) & \darkred{``Guns \textbf{are like} hammers—they’re both tools with metal parts that could be used to kill someone. And \textbf{yet it would be ridiculous} to restrict the purchase of hammers, \textbf{so} restrictions on purchasing guns are  \textbf{equally ridiculous}.''}
\\ \hline 
\textbf{Intentional \newline Fallacy} \newline (5.84\%) & ``\textbf{No one has} ever been able to prove that extraterrestrials exist, \textbf{so they must} not be real.''
\newline
\darkred{``\textbf{It's common sense} that if you smack your children, they will stop the bad behavior. \textbf{So don't tell me} not to hit my kids.''}
\\ \hline
\textbf{Fallacy of \newline Extension} & ``Their support of the discussion of sexual orientation issues is dangerous: \textbf{they advocate for} the exposure of children to sexually explicit materials, \textbf{which is wrong}.''
\\
(5.76\%) & \darkred{``They say we should cut back the defense budget. \textbf{Their position is that} they want to leave our nation completely defenseless!''}
\\ \hline
\textbf{False Dilemma} & ``You're \textbf{either} for the war \textbf{or} against the troops.''
\\
(5.76\%) & `` \darkred{\textbf{I don't want to} give up my car, so \textbf{I don't think} I can support fighting climate change.''}
\\ \hline
\textbf{Fallacy of \newline Credibility} & ``My \textbf{professor}, who has \textbf{a Ph.D. in Astronomy}, once \textbf{told me} that ghosts are real. \textbf{Therefore, ghosts are real.}''
\\
(5.39\%) & \darkred{``My \textbf{minister} says the Covid vaccine will cause genetic mutations. He \textbf{has a college degree}, and is a \textbf{holy man}, \textbf{so he must be right}.''}
\\ \hline
\textbf{Equivocation}
\newline (2.00\%) 
& ``I don't see how you can say you're an \textbf{ethical person}. It's so hard to get you to do anything; your \textbf{work ethic} is so bad''\\
& \darkred{``It is \textbf{immoral to kill an innocent human} being. Fetuses are innocent human beings. Therefore, it is \textbf{immoral to kill fetuses}.''}
\\
\bottomrule
    \end{tabular}
    \caption{Examples of the 13 logical fallacy types and their distribution in the \logicGeneral{} dataset. To illustrate of the potential impact of learning logical fallacies, we select some examples with \textbf{neutral} impact that we manually identify, and some with \darkred{\textbf{potentially negative}} impact.}
    \label{tab:data_ex}
\end{table*}

To address this pressing need and encourage more work to detect reasoning flaws, we construct a dataset of logical fallacies consisting of general logical fallacies (\textbf{\logicGeneral{}}), and a challenging extrapolation set of climate claims (\textbf{\logicClimate{}}), as shown in \cref{fig:intro}.
% \mrinmaya{Logic-aware -> structure-aware or pattern-aware?}
We find that this task
% e task of logical fallacy detection
is challenging for 12 pretrained large language models, whose performances range from 8.62\% to 53.31\% micro $F_1$ scores on the \logicGeneral{} dataset. 

By analyzing our collected dataset, we identify that logical fallacies often rely on certain false patterns of reasoning.
For example, a typical pattern in \textit{false causality} in \cref{fig:intro} is ``$\alpha$ co-occurs with $\beta$ $\Rightarrow$ $\alpha$ causes $\beta$.''
%It does not matter what \textit{A} and \textit{B} are in this case, but it is important to detect all paraphrases of \textit{A} and \textit{B}, and distill the structure of the argument.
Motivated by this, 
%As language models are not able to identify logical fallacies well,
we develop an approach to encourage language models to identify these underlying patterns behind the fallacies. 
% a structure distillation process
% \mrinmaya{Motivate why we need to think of this problem in a structured/logical way. Describe how it leads to your new approach and describe the high-level view.}
In particular, we design a \textbf{structure-aware model} which identifies text spans that are semantically similar to each other, masks them out, and then feeds the masked text instances to a classifier. This structure distillation process can be implemented atop any pretrained language model. 
% that distills the language structure from the content, and 
Experiments show that our model outperforms the best pretrained language model by 5.46\% on \logicGeneral{}, and 4.51\% on \logicClimate{}.
% . Finally, we show that models have limited performance on the more challenging test set, \logicClimate{}, but the structure-aware classifier still improves by 4.51\% over its corresponding language model baseline.

% transferring from \logicGeneral{} helps the prediction on \logicClimate{}, but still have limited extrapolation performance,
% which indicates the large improvement space for future work to capture the reasoning structure of text independent from the content.

% \mrinmaya{The contributions seem to repeat the above. Can be skipped to save space and we can end with a line that we encourage future work to explore this task}
In summary, this paper makes the following contributions:
% \noindent 1. We propose a new task of logical fallacy classification.

% \noindent 2. We collect a dataset of 2,449 samples of 13 logical fallacy types, with an additional challenge set of 1,109 climate change claims with logical fallacies.

% \noindent 3. We conduct extensive experiments using 12 existing language models and show that these models have very limited performance on detecting logical fallacies.

% \noindent 4. We design a structure-aware classifier as a baseline model for this task, outperforming the best language model by 5.46\%. This baseline model can also achieve 27.23\% when tested on the challenge set, which is 4.51\% higher than the language model baseline.

% Below shows bad spacing in overleaf:
\begin{enumerate}[topsep=0.1ex,itemsep=-1ex]
    \item We propose a new task of logical fallacy classification.
    \item We collect a dataset of 2,449 samples of 13 logical fallacy types, with an additional challenge set of 1,109 climate change claims with logical fallacies.
    \item We conduct extensive experiments using 12 existing language models and show that these models have very limited performance on detecting logical fallacies.
    \item We design a structure-aware classifier as a baseline model for this task, which outperforms the best language model.
    % by 5.46\% on \logicGeneral{} and 4.51\% on \logicClimate{}.
    % This structure-aware baseline can also achieve 27.23\% when tested on the challenge set, which is 4.51\% higher than the best language model.
    \item We encourage future work to explore this task and enable NLP models to discover erroneous patterns of reasoning.
\end{enumerate}

\section{Logical Fallacy Dataset}
First, we introduce our data. Our logical fallacy dataset consists of two parts: a) a set of common logical fallacies (\logicGeneral{}), and b) an additional challenge set of logically fallacious claims about climate change (\logicClimate{}).
\subsection{Common Logical Fallacies: \logicGeneral{}}
\paragraph{Data Collection}
The \logicGeneral{} dataset consists of common logical fallacy examples collected
from various online educational materials meant to teach or test the understanding of logical fallacies among students. 
%Specifically, 
We automatically crawled examples of logical fallacies from three student quiz websites, \href{https://quizizz.com/admin/search/logical\%20fallacies?sortBy=\_score\&grade=all\&subject=All\&langs=English&numQuestions=\&duplicates=false\&studentQuizzes=false\&safeSearch=true\&type=quiz}{{Quizziz}}, \href{https://study.com/learn/fallacy-questions-and-answers.html}{study.com} and \href{https://www.proprofs.com/quiz-school/topic/logical-fallacy}{ProProfs}
(resulting in around 1.7K samples),
% 52704 + 137 + 40
% (resulting in 52K raw, unfiltered data samples), 
and manually collected fallacy examples from some additional websites
%teaching and explaining logical fallacies 
recommended by Google search (resulting in around 600 samples). 
% manuall collection: 400 by Zhijing, 118 by Jad, 127 by Safiyyah.
More data collection and filtering details are in \cref{appd:logic_general_filtering}.
\begin{table}[h]
    \centering
    %\small
      \resizebox{\columnwidth}{!}{%
    \begin{tabular}{lccccc}
    \toprule
     & \textbf{\# Samples} & \textbf{\# Sents} & \# \textbf{Tokens} & \textbf{Vocab}  \\ \hline
\textbf{Total Data} & 2,449 & 4,934 & 71,060 & 7,624 \\
\quad Train & 1,849 & 3,687 & 53,475 & 6,634 \\
\quad Dev & 300 & 638 & 8,690 & 2,128 \\
\quad Test & 300 & 609 & 8,895 & 2,184 \\
    \bottomrule
    \end{tabular}
    }
    \caption{Statistics of the \logicGeneral{} dataset.}
    \label{tab:edu_stats}
\end{table}

The entire \logicGeneral{} dataset contains 2,449 logical fallacy instances across 13 logical fallacy types.  We randomly split the data into train, dev, and test sets; dataset statistics are shown in \cref{tab:edu_stats}, and the distribution and examples of each type 
% in \cref{tab:sent_label_list}. We also list some examples of the five most common fallacies 
in \cref{tab:data_ex}. More details of each fallacy type are in \cref{appd:logic_types}.

% \begin{table}[]
%     \centering\small
%     \begin{tabular}{lcc}
% \toprule
% \textbf{Logical Fallacy Type} & \textbf{Frequency in Data} \\ \hline
% Faulty Generalization & 18.01 \% \\
% Ad Hominem & 12.33 \% \\
% Ad Populum & 9.47 \% \\
% False Causality & 8.82 \% \\
% Circular Claim & 6.98 \% \\
% Appeal to Emotion & 6.82 \% \\
% Fallacy of Relevance & 6.61 \% \\
% Deductive Fallacy & 6.21 \% \\
% Intentional & 5.84 \% \\
% Fallacy of Extension & 5.76 \% \\
% False Dilemma & 5.76 \% \\
% Fallacy of Credibility & 5.39 \% \\
% Equivocation & 2.00 \% \\
% % If we let the government ban SUVs, then they're going to want to ban trucks! Then they'll want to ban sports cars! Then they'll want to ban all cars!
% % There is a lot of commotion regarding saving the environment. We simply cannot make this world into the Garden of Eden. Pursuing perfection is impossible and pointless. And besides, even Adam and Eve got bored in Eden!
% % A book argues that global warming is not actually happening, and cites the research of one environmental scientist who has been studying climate change for several years.
%     \bottomrule
%     \end{tabular}
%     \caption{The 13 logical fallacy types in our \logicGeneral{} dataset.}
%     \label{tab:sent_label_list}
% \end{table}

\myparagraph{Comparison with Existing Datasets}
Due to the challenges of data collection, all previous existing datasets on argument quality are of limited size. 
% We obtained a dataset of 2,449 logical fallacy samples in total. 
In \cref{tab:compare}, we draw a comparison among our dataset and two existing datasets: an argument sufficiency classification dataset \cite{stab-gurevych-2017-recognizing}, which proposes a binary classification task to identify whether the evidence can sufficiently support an argument, and another dataset dedicated for a specific type of logical fallacy called \textit{ad hominem}, or \textit{name-calling} \cite{habernal-etal-2018-name} 
%which detects fallacies 
where the arguer attacks the person instead of the claim.

\begin{table}[ht]
%\small
  \resizebox{\columnwidth}{!}{%
    \centering
    \setlength\tabcolsep{3pt}
    \setlength\extrarowheight{2pt}
    \begin{tabular}{lcclcc}
    \toprule
    \textbf{Dataset} & \textbf{\# Claims }
    % & Sents & Tokens 
    & \textbf{\# Classes} & \multicolumn{1}{c}{ \textbf{Purpose}} \\ \hline
    Arg. Suff. & 1,029 
    % & 4,593 & 97,370
    & Binary & Detect insufficiency
    \\
    Ad Homi. & 2,085 
    % & 3,866 & -- 
    & Binary & Detect name calling
    \\
    \textbf{\logicGeneral{}} & \textbf{2,449 }
    % & 4,302 & 66,926
    & \textbf{Multiple} & Detect \textbf{all} fallacy types
    \\
    \bottomrule
    \end{tabular}
    }
    \caption{Comparison of our logical fallacy dataset with two existing datasets, argument sufficiency classification \cite{stab-gurevych-2017-recognizing} and ad hominem classification \cite{habernal-etal-2018-name}.}
    \label{tab:compare}
\end{table}
Compared to the existing datasets, our dataset has two advantages: (1) we have a larger number of claims in our dataset, and (2) our task serves the more general purpose of detecting all fallacy types instead of a single fallacy type. These two characteristics  make our dataset significantly more challenging.
%as it involves multiple categories of logical fallacies.
%\mrinmayaside{The above is not clear. Do you want to say that your dataset is multi-label? Or do you want to say that you have more than one logical fallacy type}

%\mrinmayaide{Ah! The below can be argued. Also, now that I think of it, should our task be called fallacy detection?}
%Note that our dataset size could be larger if we were to further obtain data of logically sound claims as a negative class, but we do not do this in case the models learn to capture artifacts or spurious correlations that co-occur with the logically sound claims collected from a different source.

\subsection{Challenge Set: \logicClimate{}}\label{sec:doc_data}

Logical fallacy detection on climate change is a small step towards promoting consensus and joint efforts to fight climate change. We are interested in whether models learned on the \logicGeneral{} dataset can generalize well to real-world discussions on climate change. Hence, we collect an extrapolation set \logicClimate{} which consists of all climate change news articles from the Climate Feedback website\footnote{\url{https://climatefeedback.org/feedbacks/}} by October 2021.

For each news article, we ask two different annotators who are native English speakers to go through each sentence in the article, and label all logical fallacies if applicable. Since directly classifying the logical fallacies at the article level is too challenging, we let the annotators select the text span while labeling the logical fallacies, and we compose each sample using the sentence containing the selected text span as logical fallacies. Details of the annotation process are described in \cref{appd:collection_logicclimate}.
%, as well as its context, for which we heuristically extract the sentence immediately before and the sentence immediately after the selected sentence.
%\mrinmaya{Why do we need sentence before and after?}
%\abhinav{correction needed,we are no longer using the sentence before and after, in the final implementation}
\begin{figure*}[t]
    \centering
    \includegraphics[width=\textwidth]{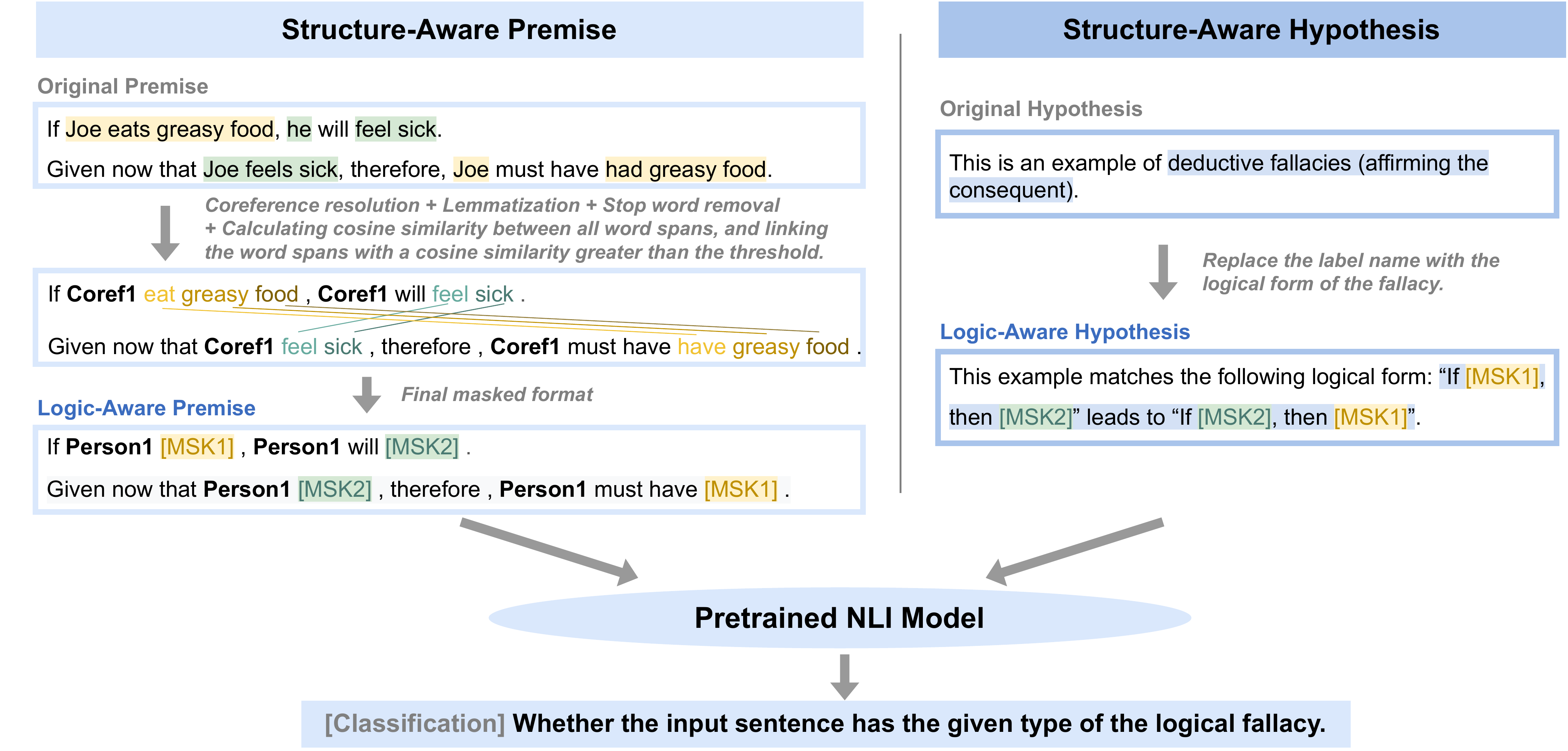}
    \caption{Our baseline model is a structure-aware classifier based on pretrained NLI model, with a structure-aware premise and structure-aware hypothesis. The structure-aware premise masks the content words to distill the argument structure. Specifically, we first resolve the coreferences, and then apply Sentence-BERT to match the lemmatized word spans (excluding the stopwords) whose contextualized embeddings have a cosine similarity larger than a certain threshold. And the structure-aware hypothesis uses the standard logical form of the given fallacy type.
    }
    \label{fig:model}
\end{figure*}

% Each news article is labeled with its corresponding logical fallacy types compiled by an editor according to the suggestions by multiple expert reviewers such as professors, senior scientists and other researchers. 
% For example, there can be articles with comments compiled from eight experts. 
\begin{table}[t]
    \centering\small
    \begin{tabular}{lc}
\toprule
\textbf{Logical Fallacy Type} & \textbf{Frequency in Data} \\ \hline
Intentional Fallacy & 25.58\% \\
Appeal to Emotion & 11.37\% \\
Faulty Generalization & 10.18\% \\
Fallacy of Credibility & 9.90\% \\
Ad Hominem & 7.84\% \\
Fallacy of Relevance & 7.80\% \\
Deductive Fallacy & 6.50\% \\
False Causality & 5.11\% \\
Fallacy of Extension & 4.91\% \\
Ad Populum & 4.55\% \\
False Dilemma & 3.80\% \\
Equivocation & 1.94\% \\
Circular Claim & 0.51\% \\
\bottomrule
    \end{tabular}
    \caption{Logical fallacy types and their frequencies in the \logicClimate{} dataset.}
    \label{tab:label_freq}
\end{table}
In total, the \logicClimate{} dataset has 1,079 samples of logical fallacies with on average 35.98 tokens per sample, and a vocabulary of 5.8K words. The label distributions are in \cref{tab:label_freq}. We provide examples of each fallacy in \logicClimate{} in \cref{appd:data_climate_examples}.

% We randomly split the data into train, dev, and test sets, and show the dataset statistics in \cref{tab:climate_stats}. 
% \begin{table}[t]
%     \centering
%     \small
%     \begin{tabular}{lccccc}
%     \toprule
%      & \textbf{\# Samples} & \textbf{\# Sents} & \textbf{\# Tokens} & \textbf{Vocab}  \\ \hline
% \textbf{Total Data} &  1,109 & 5,455 & 145,622 & 8,691 \\
% \quad Train & 914 & 3,598 & 99,831 & 7,013 \\
% \quad Dev & 260 & 1,092 & 27,913 & 3,098 \\
% \quad Test & 177 & 765 & 17,878 & 2,767 \\
%     \bottomrule
%     \end{tabular}
%     \caption{Statistics of the \logicClimate{} dataset.}
%     \label{tab:climate_stats}
% \end{table}

\section{A Structure-Aware Model}
The task of logical fallacy classification is 
%distinct from many established NLP tasks such as sentiment classification 
unique in that logical fallacies are not just about the content words (such as the sentiment-carrying words in a sentiment classification task), but more about the ``form'' or ``structure'' of the argument. 

To advance the ability of models to detect fallacious logical structures, we draw inspirations from the history of logic \cite{russell2013history}. If we look into the time when Aristotle made his attempt to formulate a systematic study of logical, one of the most notable advancements is to move from contents to symbols, based on which Aristotle developed a system of rules \cite{gabbay2004handbook}. For example, he uses $\alpha$, $\beta$, $\gamma$ to distill arguments such as ``Socrates is a man. All men are mortal. Therefore Socrates is mortal.'' into forms such as ``$\alpha$ is a $\beta$. All $\beta$ are $\gamma$. Therefore, $\alpha$ is $\gamma$.'', where the variables act as placeholders. After establishing a system of valid and invalid argument structures, philosophers can refute a fallacious argument by comparing it to a list of fallacious logical forms \cite{aristotle2006sophistical}.

Based on such inspirations, we propose a structure-aware classifier as a baseline model for our logical fallacy detection task.
%that serve to motivate future work on this task. 
%Specifically, w
We first introduce a commonly used classification framework using pretrained models on natural language inference (NLI) in \cref{sec:backbone}, and then we will propose our structure distillation process in \cref{sec:logic_designs}.

% We propose a baseline model that distills the language structure from content and incorporate this design to 
% the commonly used classification pipeline of pretrained models on natural language inference (NLI) (briefly introduced in \cref{sec:backbone}). 
% We will first briefly introduce the backbone of the model based on, NLI-based language models, in \cref{sec:backbone}, and then propose our structure-aware designs in \cref{sec:logic_designs}.

\subsection{Backbone: NLI-Based Classification with Pretrained Models}\label{sec:backbone}
Motivated by the success of adapting NLI for classification tasks with unseen labels \cite{yin-etal-2019-benchmarking}, we choose pretrained language models on NLI as the backbone of our logical fallacy classifier. 

Specifically, a standard NLI-based pretrained language model for classification takes the sentence to classify as the \textit{premise}. Then the model composes a \textit{hypothesis} using the template of ``This example is \texttt{[label name]}.'' The classifier checks whether the premise can entail the hypothesis.
% For multi-class classification, this NLI process is repeated for each class label. If the task is single-label multi-class classification (i.e., one ground-truth label for each sample), then the class label with the largest probability that the model predicts that the premise \textit{entails} the hypothesis is chosen, whereas if the task is multi-label multi-class classification (i.e., multiple ground-truth label for each sample), then a threshold is chosen, and all the classes that have the entailment probability larger than the threshold is predicted \cite{yin-etal-2019-benchmarking}.
This NLI framework makes it easy for pretrained language models to adapt to unseen class labels such as our logical fallacy types.
% An additional benefit of using NLI-based classifiers is that they can be extended to unseen logical fallacy types and serve as a zero-shot classifier for new use cases.

\subsection{Distilling Structure from Content}\label{sec:logic_designs}
% The task of logical fallacy classification is distinct from many established NLP tasks such as sentiment classification in that logical fallacies are not just about the content words (such as the sentiment-carrying words in the sentiment classification task), but more about the ``form,'' or the argument structure of language. 
% It is challenging for the model to learn content-agnostic argument structures. 

To build a model that encourages more attention to the \textit{structure}
%rather than the \textit{content} 
of the text, we modify the premise and the hypothesis provided to the backbone NLI model (as shown in \cref{fig:model}): called the \textit{structure-aware premise} and \textit{structure-aware hypothesis}.

\myparagraph{Structure-Aware Premise} 
% We make some adaptations of the standard NLI models for classification that takes as input the text as the premise, and the class label as the hypothesis. 
Inspired by the process how ancient Greek philosophers refute an argument they have heard, we design an argument structure distiller by masking out content words in the premise (i.e., input text) and outputting a logical form with placeholders.
In the example in \cref{fig:model}, ``Jack is \textit{a} good athlete. Jack {comes from} Canada. 
Therefore, \textit{all} Canadians {are} good athletes.'', 
we want the model to pay more attention to the structure as opposed to contents such as ``good athletes.'' Thus, we build a distilled argument with placeholders ``[MSK1] is a [MSK2]. [MSK1] comes from [MSK3]. Therefore, all [MSK3] are [MSK2].''

% \mrinmaya{Remove this para?}
% To put it straightforward, it does not matter whether Jack is a good or bad athlete, or whether he is an athlete or an actor, 
% but it is about the fact that ``good athlete'' appears twice in the claim, and is used to describe different entities.
% but it is the structure that is problematic, namely the fallacious ``A has attribute B. A is a subset of C. $\Rightarrow$ All C has attribute B.''
% , ``\textit{If }Joe eats greasy food, he \textit{will }feel sick. \textit{Given now that }Joe feels sick, \textit{therefore, }Joe \textit{must have }had greasy food.'', it does not matter whether Joe is feeling sick or feeling good, or whether it is about eating greasy food or raw food, but it is the structure that is problematic, namely the fallacious ``If A, then B. $\Rightarrow$ If B, then A.''
% Here, it is not important whether the text is about ``good athlete''
% % greasy food''
% or not, 
% one to describe Jack, and the other to describe all Canadians.

As shown in \cref{fig:model}, to distill the premise into the logical form, we identify all text spans that are paraphrases of each other and replace them with the same mask. Specifically, we first conduct coreference resolution using the \texttt{CoreNLP} package \cite{manning-etal-2014-stanford}. Then, to identify word spans that are paraphrases of each other, we consider only non-stop words, lemmatize them via the \texttt{Stanza} package \cite{qi-etal-2020-stanza}, and 
% we transform the original text input by first doing coreference resolution using the \texttt{CoreNLP} package \cite{manning-etal-2014-stanford}, and then preprocessing the text by lemmatization and stop word removal via the \texttt{Stanza} package \cite{qi-etal-2020-stanza}. Based on the preprocessed text input, we design some simple heuristics to identify text spans that are paraphrases of each other. Specifically, we
represent each word by its contextualized embedding generated by Sentence-BERT \cite{reimers-gurevych-2019-sentence}, and calculate pair-wise cosine similarity. When the cosine similarity is larger than a threshold (by 
% ,\footnote{We tune the threshold using 
a grid search on the dev set),
% .} 
we identify the two words as similar. For illustration, we create a link between similar word pairs in \cref{fig:model}. When there are contiguous sequences of words that are linked to each other (e.g., ``good athlete'' and ``good athletes''), we merge them and end up with two multi-word spans that are similar to each other. For each group $i$ of similar text spans, we replace them with a mask token $\texttt{[MSK}_i\texttt{]}$.
%\mrinmaya{How is this threshold decided?}
%\abhinav{It is decided based on examples in the dev set, and manually seeing what works best}

\myparagraph{Structure-Aware Hypothesis} NLI-based classification models \citep{yin-etal-2019-benchmarking} typically compose the hypothesis as a template sentence ``This is an example of \texttt{[label name]}.'' 
%\mrinmaya{I think you are not talking about standard NLI models here. I imagine you are talking about the NLI based classification approach by Yin}
However, in order to help our model perform a structure-aware matching of the logical fallacy instance, we also augment the hypothesis with the logical form for the logical fallacy type. For example, the logical form for \textit{faulty generalization} in the example in \cref{fig:model} is changed to:
``[MSK1] has attribute [MSK2]. [MSK1] is a subset of [MSK3].
Therefore, all [MSK3] has attribute [MSK2].''
% ``If A happens, then B will happen. Given now that B happens, therefore, A must have happened.''

To look up the logical form of each fallacy, we refer to websites that introduce the logical fallacies, extract the expressions such as ``Circular reasoning is often of the form: `A is true because B is true; B is true because A is true.''', and compile the logical forms using our masking format.
We provide the list of logical forms in \cref{appd:logic_types}.
% To make the hypothesis consistent with the masked premise, we also adopt the mask token $\texttt{[MSK}_i\texttt{]}$ in the standard logical forms, as shown in the ``structure-aware hypothesis'' part of \cref{fig:model}.
% \mrinmaya{Need to describe the list of logical forms. Human written? How many per class, etc... here}

\section{Experiments}
\subsection{Experimental Setup}
% On our datasets \logicGeneral{} and \logicClimate{}, we suggest 
\myparagraph{Evaluation Metrics} Since the nature of the logical fallacy detection task is a multi-label classification with class imbalance, we use \textit{micro $F_1$} as the main evaluation metric. Additionally, we also report precision, recall and accuracy.

\myparagraph{Baselines} We test the performance of 12 existing large language models, including five zero-shot models and seven finetuned models. For zero-shot models, we use the zero-shot classifier by \texttt{transformers} Python package \cite{wolf-etal-2020-transformers} implemented using RoBERTa large \cite{liu2019roberta} and BART large \cite{lewis-etal-2020-bart} finetuned on the multi-genre natural language inference (MNLI) task \cite{williams2018broad}. We also include the task-aware representation of sentences (TARS) \cite{halder2020task} provided by FLAIR \cite{akbik2019flair}.
% Apart from the general-purpose zero-shot classifiers, 
Moreover, we also try directly using GPT-2 \cite{radford2019language} and GPT-3 \cite{brown2020language}. For GPT-3, we designed a prompt for the auto-completion function to predict the label of text, and for GPT-2, we calculate the perplexity of every possible label with the text and choose the label with the lowest perplexity. 
% As the last baseline, we also obtain the results of random guessing. 
See Appendix~\ref{appd:zero} for more implementation details. 

For finetuned baselines, we finetune seven
commonly used pretrained language models on the \logicGeneral{} dataset, including 
ALBERT~\cite{lan2020albert}, BERT~\cite{devlin2019bert}, BigBird~\cite{zaheer2020big}, DeBERTa~\cite{he2021deberta}, DistilBERT~\cite{sanh2019distilbert}, \electra{}~\cite{clark2020electra}, MobileBERT~\cite{sun2020mobilebert}, and RoBERTa~\cite{liu2019roberta}. 
% Before doing inference on the test set, we
See Appendix~\ref{appd:finetune_exp} for implementation details.

\myparagraph{Implementation Details}
We describe the implementation details of our structure-aware classifier in \cref{appd:implementation_our_model}.
% \mrinmaya{Any details on model training to be discussed before results? Possibly point to the appendix?}
% \abhinav{A correction is needed here, no threshold is used, rather we use the argmax of the 3 classes (Entailment, Neutral, Contradiction)}
\subsection{Main Results}\label{sec:finetuning}
We test how well existing language models can address the task of logical fallacy classification, and check whether our proposed model can lead to performance improvement. 
% In \cref{tab:fewshot}, we first report the performance of 12 pretrained language models on our \logicGeneral{} dataset, including five zero-shot models and seven finetuned models. Then, we use the best-performing language model, \electra, as the backbone for our proposed structure-aware classifier.

\begin{table}[t]
\centering
  \resizebox{\columnwidth}{!}{%
    \setlength\tabcolsep{5pt}
\begin{tabular}{l  cccccc}
\toprule
{} &  $F_1$ 
% &  Ma$F_1$ 
& P & R & Acc \\
\midrule
% Majority &  9.92 &  0.20 & 10.00 & 9.84 &  9.33 \\
Random & 12.02 
% & 10.96 
& 7.24 & 35.00 & 0.00\\ \midrule
\multicolumn{5}{l}{\textbf{\textit{Zero-shot classifiers directly tested on \logicGeneral{}}}}
\\
TARS & 8.62 
% & 5.84 
& 3.86 & 6.67 & 2.33 \\
BART-MNLI & 11.05 
% & 9.79 
& 6.63 & 33.67 & 0.00 \\
GPT3 & 12.20 
% & 8.86 
& 12.00 & 12.00 & 12.00 \\
RoBERTa-MNLI & 12.22 
% & 11.51 
& {7.51} & 36.00 & 0.33 \\
GPT2 & {13.67} 
% & 1.71 
& {13.67} & {13.67} & {13.67} \\
\midrule
\multicolumn{5}{l}{\textbf{\textit{Finetuned and tested on \logicGeneral{}}}}
\\
ALBERT  & 12.50  
% & 12.10  
& 6.67  & {100.00}  & 0.00 \\
BigBird  & 15.02  
% & 11.58  
& 8.61  & 90.00  & 0.33 \\
DistilBERT & 26.96 
% & 26.12 
& 22.06 & 74.00 & 4.67 \\
MobileBERT  & 35.68  
% & 28.62  
& 29.05  & 71.00  & 7.33 \\
BERT  & 45.80  
% & 36.59  
& 40.73  & 73.67  & 18.00 \\
DeBERTa  & 50.29  
% & 40.40  
& 45.79  & 73.00  & 24.67 \\
\electra{}  & 53.31  
% & {50.21}  
& 51.59  & 72.33  & 35.66 \\
\ourmodel{} & \textbf{58.77}
% & 48.91  
& {55.25}  &63.67  & {47.67} \\
\midrule
\multicolumn{5}{l}{\textbf{\textit{Ablation study on the proposed model}}}
\\
Raw Prem. + Str. Hypo. & 56.72  
% & 46.71  
& 54.87  & 76.67  & 37.67 \\
Str. Prem. + Raw Hypo. & 44.56 &	39.74	& 71.00 &	18.33
\\
\bottomrule
\end{tabular}
}
\caption{Model performance on \logicGeneral{} by the ascending order of the main metric, micro $F_1$ ($F_1$).
% We report all metrics including $F_1$, 
% macro $F_1$ (Ma$F_1$), 
In addition, we also report the
precision (P), recall (R), and accuracy (Acc). In the ablation study, we report the performance of two settings: a raw premise (i.e., keeping the original text input) with a structure-aware hypothesis (Raw Prem. + Str. Hypo.), and a structure-aware premise with a raw hypothesis (Str. Prem. + Raw Hypo.).}
\label{tab:fewshot}
\end{table}
\myparagraph{Zero-Shot Classifiers}
In \cref{tab:fewshot}, we first look into some commonly used off-the-shelf zero-shot classification models. 
Surprisingly, most zero-shot classifiers are not much better than randomly choosing a label (i.e., the ``Random'' baseline in \cref{tab:fewshot}). The RoBERTa-MNLI classifier and GPT2, which achieve 12.22\% and 13.67\% $F_1$ scores, respectively are only marginally better than random guessing.

% The best-performing ones are BART and GPT3, which achieve 14.24\% and 13.99\% micro $F_1$ scores, respectively. Their macro $F_1$ scores (5.87\% and 4.67\%) are much lower than the micro $F_1$ scores because the models mostly predict correctly for common classes, but do not do well on uncommon classes. A possible reason is that LLMs have not seen many logical fallacy types and do not have a good understanding of what expressions correspond to what logical fallacies.

\myparagraph{Finetuned Models}
We further look into the effectiveness of finetuned large language models.
The model performance is shown in ascending order in \cref{tab:fewshot}. According to our main metric, $F_1$, the best language model is \electra{}, which achieves 53.31\% $F_1$ scores, followed by DeBERTa which achieves 50.29\%.

Then, we adopt \electra{} as the backbone model to test our proposed structure-aware classifier (denoted as \ourmodel{}). Our model outperforms \electra{} by 5.46\%, which is a fairly large margin. This implies the importance of encouraging the model to shift its attention to the logical form. Our model also achieves the highest exact match result, 47.67\%, which is 12.01\% better than the best performance among all language models finetuned in the standard way.

\myparagraph{Ablation Study}
Through the ablation study in \cref{tab:fewshot}, we can see that raw premise (i.e., keeping the original text input) with structure-aware hypothesis yields 56.72\%, which can be attributed to the fact that the logical form provides richer information than just the label name. On the contrary, the structure-aware premise with the raw hypothesis of just the label name leads to a much worse result, perhaps because the model cannot easily figure out the correspondence between the masked text input and the label name. The ablation study also demonstrates that the best performance of our model comes from the matching between the logical form and the masked text input.
% and show that both the structure-aware premise and hypothesis are important to the model, and the latter is more important. \mrinmaya{add what we can conclude by this? is this a robustness test? the fact that w/o premise works well is quite interesting.}
% \abhinav{I think we can conclude that the LogicalForm is the most important part for the prediction, and that the content-masking is just a heuristic that helps it match to the Logical Form a bit better}
% In the ablation study in \cref{tab:fewshot}, we also check how much the structure-aware premise and structure-aware hypothesis contribute to the overall performance. For simplicity, if we focus on the $F_1$ scores, the vanilla NLI gives 53, only transforming the premise but not hypothesis gives 44, only transforming the hypothesis but not the premise gives 56, and transforming both the hypothesis and the premise gives 58.

% The precision and recall values in \cref{tab:fewshot} show an interesting trend. Most models can achieve a high recall, with the highest recall 80.98 achieved by RoBERTa. Recall that multi-label classification is achieved by letting the model judge the possibility of each logical fallacy class independently, and we use a default threshold of 0.5 to judge whether a class should be judged as a predicted class.

\subsection{Class-Specific Performance}\label{sec:exp_subtype}
In addition to the overall performance of our proposed \ourmodel{} model, 
% we gain a better picture of the results by 
we further analyze its class-specific performance in \cref{tab:electra}.
% , we list the class-related performance of the best-performing model \electra{} for the 12 most common classes in the test data.
\begin{table}[t]
\centering
  \resizebox{\columnwidth}{!}{%
\setlength\tabcolsep{5pt}
\setlength\extrarowheight{2pt}
\begin{tabular}{lcccc}
\toprule
{}  &  $F_1$ &     P &      R & Freq.
% Test \% 
\\
\midrule
% Faulty Generalization & 60.24 & 47.62 & 81.97 & 20.33 \\
% Ad Hominem & 78.65 & 72.92 & 85.37 & 13.67 \\
% Ad Populum & 79.45 & 67.44 & 96.67 & 10.00 \\
% Fallacy of Relevance & 39.22 & 37.04 & 41.67 & 8.00 \\
% Appeal To Emotion & 50.00 & 48.00 & 52.17 & 7.67 \\
% Fallacy of Extension & 49.18 & 37.50 & 71.43 & 7.00 \\
% Circular Claim & 46.43 & 35.14 & 68.42 & 6.33 \\
% False Causality & 58.82 & 62.50 & 55.56 & 6.00 \\
% Fallacy of Credibility & 58.82 & 58.82 & 58.82 & 5.67 \\
% Intentional & 26.23 & 17.39 & 53.33 & 5.00 \\
% Fallacy of Logic & 25.81 & 16.67 & 57.14 & 4.67 \\
% False Dilemma & 55.00 & 39.29 & 91.67 & 4.00 \\
% Equivocation & 33.33 & 100.00 & 20.00 & 1.67 \\

Faulty Generalization & 60.24 & 47.62 & 81.97 & 18.01 \\
Ad Hominem & 78.65 & 72.92 & 85.37 & 12.33 \\
Ad Populum & 79.45 & 67.44 & 96.67 & 9.47 \\
False Causality & 58.82 & 62.50 & 55.56 & 8.82 \\
Circular Claim & 46.43 & 35.14 & 68.42 & 6.98 \\
Appeal to Emotion & 50.00 & 48.00 & 52.17 & 6.82 \\
Fallacy of Relevance & 39.22 & 37.04 & 41.67 & 6.61 \\
Deductive Fallacy & 25.81 & 16.67 & 57.14 & 6.21 \\
Intentional Fallacy & 26.23 & 17.39 & 53.33 & 5.84 \\
Fallacy of Extension & 49.18 & 37.50 & 71.43 & 5.76 \\
False Dilemma & 55.00 & 39.29 & 91.67 & 5.76 \\
Fallacy of Credibility & 58.82 & 58.82 & 58.82 & 5.39 \\
Equivocation & 33.33 & 100.00 & 20.00 & 2.00 \\
\hline
Overall & {58.77}  
% & 48.91  
& {55.25}  &63.67  & {100} \\
\bottomrule
\end{tabular}}
\caption{Class-specific performance achieved by \ourmodel{}. For each class, we report the $F_1$ score, precision (P), recall (R), and the frequency (Freq.) of the class in the \logicGeneral{} dataset. Note that the Freq. column is copied from \cref{tab:data_ex}.}
\label{tab:electra}
\end{table}

Many of the logical fallacy classes can reach $F_1$ scores close to the overall $F_1$ of 58.77\%. 
However, there are some logical fallacy types with relatively higher or lower performance. As the prediction performance can depend on both the difficulty of identifying a logical fallacy type as well as the number of training samples for that type, we also provide the frequency (\%) of each logical fallacy in \cref{tab:electra}.

We can notice that the best-performing classes are \textit{ad populum} ($F_1$=79.45\%) and \textit{ad hominem} ($F_1$=78.65\%), which even outperform the most frequent class, \textit{faulty generalization} ($F_1$=60.24\%). A possible reason can be that \textit{ad populum} can be detected often when there are numbers or terms that refer to a majority of people, and \textit{ad hominem} uses insulting words or undermines the credibility of a person.

We further look into logical fallacies that are difficult to learn. For example, among the four logical fallacies with a similar frequency of 6+\% in the dataset, namely \textit{circular claim}, \textit{appeal to emotion}, \textit{fallacy of relevance}, and \textit{deductive fallacy}, the one that is the most difficult to learn is \textit{deductive fallacy} ($F_1$=25.81\%), which has the lowest $F_1$ across all 13 classes. This might be a combined effect of the difficulty of distilling the formal logic from various content words in this case, and also that there can be several more forms of deductive fallacies which are not covered by our approach. This could be an interesting direction for future work.

% \red{Future work: Confusion matrix between false causality and hasty generalization. Explain the high recall of ad populum, red herring, etc.}

\subsection{Extrapolating to \logicClimate{}}
\begin{table}[t]
\centering
\small
%   \resizebox{\columnwidth}{!}{%
    % \setlength\tabcolsep{3pt}
\begin{tabular}{l  cccccc}
\toprule
{} &  $F_1$ 
% &  Ma$F_1$ 
& P & R
% & ExMa 
\\
\midrule
% % Majority &  9.92 &  0.20 & 10.00 & 9.84 &  9.33 \\
\multicolumn{4}{l}{\textbf{\textit{Direct Transfer}
% (Directly testing 
% \textit{Finetuned on \logicGeneral{} and tested
% on \logicClimate{})
}
}
\\
% ALBERT  \\
% BigBird \\
% DistilBERT & 19.69 & 11.79 & 55.48 & 0.36  \\
% MobileBERT & 19.83 & 12.50 & 48.14 & 0.00 \\
% BERT  & 23.30  & 15.67 & 42.62 & 1.17 \\
% DeBERTa  \\
\electra{}  & 22.72
& 18.68 & 35.85
% & 5.20 
\\
\ourmodel{} & \textbf{27.23} 
 & 20.46 & 45.12
%  & 3.34 
 \\ 

\hline
\multicolumn{4}{l}{\textbf{\textit{Finetuned further on \logicClimate{}}
}}
\\
\electra{} (Ft) & 23.71
& 20.86 & 23.09
% & 13.97 
\\
\ourmodel{} (Ft) & \textbf{29.37} 
& {17.66} & 67.22
% & 8.37 
\\
\bottomrule
\end{tabular}
% }
\caption{Performance of \textit{direct transfer} models trained on \logicGeneral{} and tested on \logicClimate{}. We also include additional results of the same two models further finetuned and tested on \logicClimate{}. Since \logicClimate{} is a multi-label classification, we omit the accuracy as it is not applicable here.}
\label{tab:res_cc_direct}
\end{table}

We also test our models on the more challenging test set, \logicClimate{}, to check how well the models can extrapolate to an unseen domain, namely claims in climate change news articles.
We use the two best-performing models trained on \logicGeneral{}, namely the best language model \electra{} and our proposed \ourmodel{} model.

% We use the same list of models finetuned on \logicGeneral{} in \cref{tab:fewshot}, and test their performance on the \logicClimate{} test set.
In \cref{tab:res_cc_direct}, the direct transfer performance is calculated by directly using the two models trained on \logicGeneral{} and testing them on the entire \logicClimate{}.
Although both 
% all 
models drop drastically when transferring to the unseen \logicClimate{} challenge set, our model \ourmodel{} achieves the higher performance, 27.23\%, and still keeps its relative improvement of 4.51\% over the \electra{} baseline.
%This echoes with the purpose of this challenge set -- models (and people) that understand the form of logical fallacies should be able to extrapolate to new scenarios and unseen facts.

We also include an additional experiment of finetuning the two models on \logicClimate{}, where both show improvements, and \ourmodel{} outperforms \electra{} by a larger margin of 5.66\%. The detailed setup of this additional experiment is in \cref{appd:logic_climate_finetune}.
As we can see, even the finetuned numbers are still lower than those of \logicGeneral{}, so we encourage more future work to enhance the out-of-domain generalizability of logical fallacy classifiers.
% \mrinmaya{Say something about the fact that these numbers are still too low in comparison with the other set. And others should work on this...}

\subsection{Error Analysis}
\begin{table}[t]
    \centering \small
    \begin{tabular}{p{0.95\columnwidth}}
    \toprule
    \multicolumn{1}{c}{\textbf{\textit{Correct Predictions}}} \\
    “You should drive on the right side of the road because that is what the law says, and the law is the law.”
    \\
    \textbf{Ground-truth label:} Circular claim
    \\ \hline
    
“Some kangaroos are tall. Some MMA fighters are tall. Therefore, some kangaroos are MMA fighters.”
\\
\textbf{Ground-truth label:} Deductive fallacy
    \\ \toprule

    \multicolumn{1}{c}{\textbf{\textit{Incorrect but Reasonable Predictions}}} \\ 
“Drivers in Richmond are terrible. Why does everyone in a big city drive like that?”
\\
\textbf{Ground-truth label:} Ad hominem
\\
\textbf{Predicted label:} Faulty generalization
\\ \hline
“Whatever happens by chance should be punished because departure from laws should be punished.”
\\
\textbf{Ground-truth label:} Equivocation
\\
\textbf{Predicted label:} Circular claim
    \\ \toprule
    \multicolumn{1}{c}{\textbf{\textit{Incorrect Predictions}}} \\
“\textbf{A} car makes less pollution than \textbf{a} bus. Therefore, car\textbf{s} are less of a pollution problem than bus\textbf{es}.”
\\
\textbf{Ground-truth label:} Faulty generalization
\\
\textbf{Predicted label:} Circular claim
\\ \hline
“Not that it ever was a thing, really. This debate -- as I argue at some length in Watermelons -- was always about \textbf{left-wing ideology, quasi-religious hysteria, and `follow the money' corruption}, never about `science.' Still, it’s always a comfort to know that `the science' is on our side too. They do so hate that fact, \textbf{the Greenies}.”
\\
\textbf{Ground-truth label:} Ad hominem and the fallacy of extension
\\
\textbf{Predicted label:} Intentional fallacy
    \\
    \bottomrule
    \end{tabular}
    \caption{Examples of correct predictions, incorrect but reasonable predictions, and incorrect predictions.}
    \label{tab:error}
\end{table}

% \mrinmaya{below do you want to say correct/incorrect prediction instead of correct/incorrect examples? also need to update figure in that case}
Next, we analyze our model predictions and common error types. We identify three categories of model predictions in \cref{tab:error}: correct predictions, incorrect but reasonable predictions, and incorrect predictions. Common among incorrect but reasonable predictions are some debatable cases where multiple logical fallacy types seem to apply, and the ground-truth label marks the most obvious one. For example, ``Drivers in Richmond are terrible. Why does everyone in a big city drive like that?'' is an example of ad hominem as it is a personal attack against drivers in Richmond, but also has some flavor of faulty generalization from ``drivers in Richmond'' to ``everyone in a big city.''

Among the incorrect predictions, we can see the difficulty of identifying the nuances in the logical forms. The sample from \logicGeneral{}, ``A car makes less pollution than a bus. Therefore, cars are less of a pollution problem than buses.'', at first glance, looks similar to circular reasoning as it seems to repeat the same argument twice. However, in fact, it is a faulty generalization from ``a car\dots a bus'' to ``cars\dots buses.''
Another sample from \logicClimate{} uses context-specific words ``left-wing ideology, quasi-religious hysteria, and `follow the money' corruption\dots the Greenies'' for ad hominem when politically criticizing climate change advocates.

\section{Limitations and Future Work}
% The main goal of this work is to propose the task of logical fallacy detection, and we use a structure-aware model to motivate future work to explore further on how language models can capture the logical structures in text.
Some limitations of the current proposed model is that it can be effective for text with clear spans of paraphrases, but does not always work for more complicated natural text, such as the journalistic style in the climate change news articles. Another limitation is that, in the scope of this work, we only explored one logical form for each fallacy type. Since there could be multiple ways to verbalize each fallacy, future work can explore if the models can match the input text to several candidate logical forms, and create a multi-way voting system to decide the most suitable logical fallacy type.

Orthogonal to model development, future work can also explore other
% Future work can potentially follow two directions: (1) technical exploration on the logical reasoning ability of NLP models on the task of logical fallacy detection, and (2) 
% socially meaningful applications
% or extensions of logical fallacy detection, 
% such as
% For the technical exploration, future work can use this dataset to develop models that are more aware of the logical and reasoning structure in language, or 
% just as a behavioral test for new language models. 
% For 
socially meaningful applications of this work, 
in line with the NLP for Social Good Initiative \cite{jin2021good,gonzalez2022nlp},\footnote{\url{https://nlp4sg.vercel.app}}
logical fallacy detection can be used in various settings: 
to validate information and help fight misinformation along with fact-checkers \cite{riedel2017simple,thorne-etal-2018-fever}, 
% to promote more critical thinking and help people uncover the chains of logic, to check argument validity as a writing assistant, 
to check whether cognitive distortions \cite{beck1963thinking,kaplan2017cognitive,lee-etal-2021-micromodels-efficient} are correlated with some types of logical fallacies,
to check whether some logical fallacies are commonly used as political devices of persuasion in politicians' social media accounts, among many other possible application cases.

\section{Related Work}

\myparagraph{Logical Fallacies} 
Logic in language is a subject that has been studied since the time of Aristotle, who considers logical fallacies as ``deceptions in disguise'' in language \citep{aristotle1991rhetoric}. Logical fallacies refer to errors in reasoning \cite{tindale2007fallacies}, and they usually happen when the premises are not relevant or sufficient to draw the conclusions \citep{johnson2006logical,damer2009attacking}.
Early studies on logical fallacies include the taxonomy \cite{greenwell2006taxonomy}, general structure of logical arguments \cite{toulmin2003uses}, and schemes of fallacies \cite{walton2008argumentation}.

% ``A cogent (or logically good) argument has individually acceptable premises that are relevant to the argument’s conclusion and, together, sufficient to draw the conclusion'' \citep{johnson2006logical,damer2009attacking}

% ``The observation that some arguments are in fact ‘deceptions in disguise’ was made already by Aristotle \citep{aristotle1991rhetoric}, for which the term fallacy has been adopted.'' 
% Taxonomy of logical fallacies in Table 6 of \citet{greenwell2006taxonomy}.
% ``\citet{toulmin2003uses} models the general structure of logical arguments, and \citet{walton2008argumentation}
% analyze schemes of fallacies and strong arguments.
% A fallacy is a kind of error that undermines reasoning \citep{tindale2007fallacies}.''

Logic is at the center of research on argumentation theory, an active research field in both the linguistics community \citep{damer2009attacking,van2013fundamentals,govier2013practical}, and NLP community \citep{wachsmuth-etal-2017-computational,wachsmuth-etal-2017-argumentation,habernal-etal-2018-argument,habernal-gurevych-2016-makes}.
% Human argumentation process and quality is widely studied since the argumentation theory (in terms of logical, rhetorical, and dialectical quality)
% % comprehensive theoretical framework on argument quality in logic and argumentation theory
% \citep{van2013fundamentals,damer2009attacking,govier2013practical}, and also in the NLP community \citep{wachsmuth-etal-2017-computational,wachsmuth-etal-2017-argumentation,habernal-etal-2018-argument,habernal-gurevych-2016-makes}. 
The most relevant NLP works include classification of argument sufficiency \cite{stab-gurevych-2017-recognizing}, ad hominem fallacies from Reddit posts \citep{habernal-etal-2018-name} and dialogs \citep{sheng-etal-2021-nice}, as well as automatic detection of logical fallacies using a rule parser \citep{nakpihautomated}. 

To the best of our knowledge, our work is the first to formulate logical fallacy classification with deep learning models, and also the first to propose logical fallacy detection for climate change news.

\myparagraph{Combating Misinformation} 
There has been an increasing trend of using NLP to combat misinformation and disinformation
% and ensure Internet freedom
\cite{emnlp-2019-natural}. Most existing works focus on fact-checking, which uses evidence to verify a claim \citep{perez-rosas-etal-2018-automatic,thorne-etal-2018-fever,riedel2017simple}.
% , which can be of quite high costs to check the claims by linking external knowledge sources. 
To alleviate the computationally expensive fact-checking procedures against external knowledge sources, some other efforts include check-worthy claim detection \cite{konstantinovskiy2028automated}, out-of-context misinformation detection \cite{aneja2021catching}, while some still need to outsource to manual efforts \cite{nakov2021automated}. We consider our work of logical fallacy detection to be independent of the topic and content, which can be an orthogonal component to existing fact-checking work. The logical fallacy checker can be used before or along with fact-checkers to reduce the number of claims to check against, by eliminating logically fallacious claims in the first place.
Logical fallacies also have some intersections with propaganda techniques \citep{da-san-martino-etal-2019-fine,da-san-martino-etal-2019-findings,da-san-martino-etal-2020-semeval,da-san-martino-etal-2020-prta}, but they are two distinct tasks, since propaganda is more about influencing people's mindsets and the means can be various types of persuasion devices, and this work on logical fallacies mainly focuses on the logical and reasoning aspect of language, with implications for enhancing the reasoning ability of NLP models.

\section{Conclusion}

This work proposed logical fallacy detection as a novel task, and constructed a dataset of common logical fallacies and a challenge set of fallacious climate claims. Using this dataset, we tested the performance of 12 existing pretrained language models, which all have limited performance when identifying logical fallacies. We further proposed a structure-aware classifier which surpasses the best language model on the dataset and the challenge set. This dataset provides a ground for future work to explore the reasoning ability of NLP models.

\section*{Acknowledgments}
We thank Kevin Jin for insightfully pinpointing the prevalence of logical fallacies in discussions of social problems. We thank the labmates at the LIT Lab at the University of Michigan, especially Ashkan Kazemi for constructive suggestions and writing advice based on existing work in fake news detection. We thank Prof Markus Leippold (University of Zürich) for insights on climate change fact verification datasets. We thank Amelia Francesca Hardy (Stanford) for discussions on pressing social problems that NLP can be promising to address.

We especially thank many annotators at the University of Michigan for helping us with the dataset, including Safiyyah Ahmed, Jad Beydoun, Elizabeth Loeher, and Brighton Pauli.
% Ahmed, Jad Beydoun, Elizabeth Loeher, and Brighton Pauli. 
Additional thanks to Jad Beydoun for helping to compile some numbers and examples in this paper. 

This material is based in part upon works supported by the German Federal Ministry of Education and Research (BMBF): Tübingen AI Center, FKZ: 01IS18039B; by the Machine Learning Cluster of Excellence, EXC number 2064/1 – Project number 390727645; by the Precision Health Initiative at the University of Michigan; by the John Templeton Foundation (grant \#61156); and by a Responsible AI grant by the Haslerstiftung.

\section*{Ethical Considerations}

The data used in this work are all from public resources, with no user privacy concerns. The potential use of this work is for combating misinformation and helping to verify climate change claims.

\section*{Contributions of the Authors}
This project was a large collaboration that would not have happened without dedicated effort from every co-author. 

The \textit{idea of the project} originated in discussions among Zhijing Jin, Bernhard Schölkopf, Rada Mihalcea, and Mrinmaya Sachan.

For the \textit{dataset collection}, Zhijing Jin led the data collection. She conducted the annotation and compilation, together with Yvonne Ding who collected the original articles for \logicClimate{}, as well as Zhiheng Lyu who automatic crawled part of \logicGeneral{}.

\textit{Analyses} of dataset characteristics and experimental results were first done by Zhijing Jin, and later updated by Abhinav Lalwani. Some analyses in the appendix were done by Zhiheng Lyu. 

For the \textit{experiments}, the first round was done by Tejas Vaidhya, the second round was done by Zhijing Jin and Xiaoyu Shen, and the final round was done by Abhinav Lalwani, including the \ourmodel{}.

\textit{Cleaning and compilation} of the code and data was done by Zhijing Jin and then updated by Abhinav Lalwani.

All co-authors contributed to \textit{writing the paper}, especially Zhijing Jin, Mrinmaya Sachan, Rada Mihalcea, Xiaoyu Shen, and Bernhard Schoelkopf.

\bibliography{references,clean_refs}
\bibliographystyle{acl_natbib}

\clearpage
\newpage
\appendix
\section{More Details of the Dataset}\label{appd:data} 
\subsection{Dataset Overview for Responsible NLP}
\myparagraph{Documentation of the artifacts:}

- Coverage of domains: general domain (e.g., educational examples of logical fallacies), and climate change news articles with logical fallacies.

- Languages: English.

- Linguistic phenomena: Logical fallacies.

- Demographic groups represented: No specific demographic groups.

\myparagraph{Annotation details:}

- Basic demographic and geographic characteristics of the annotator
population that is the source of the data: All annotators are native English speakers who are undergraduates at a university in the US.
% the University of Michigan in the US. 
There are two male annotators and two female annotators.

- How you recruited (e.g., crowdsourcing platform, students) and paid participants, and discuss if such payment is adequate given the participants’ demographic (e.g., country of residence): We broadcast the recruitment to the undergraduate CS student mailing list at a university.
% the University of Michigan. 
We received a large number of applications and selected four annotators. We followed the university's standard payment of 14 USD/hour for each student.

- How consent was obtained from annotators:
% people whose data you’re using/curating (e.g., did your instructions explain how the data would be used): 
We explained to the annotators that the data will be open-sourced for research purpose.

- Data collection protocol approved (or determined exempt) by an ethics review board: The dataset included in this work did not go through reviews by an ethics review board.

\begin{figure}[h]
    \centering
    \includegraphics[width=\columnwidth]{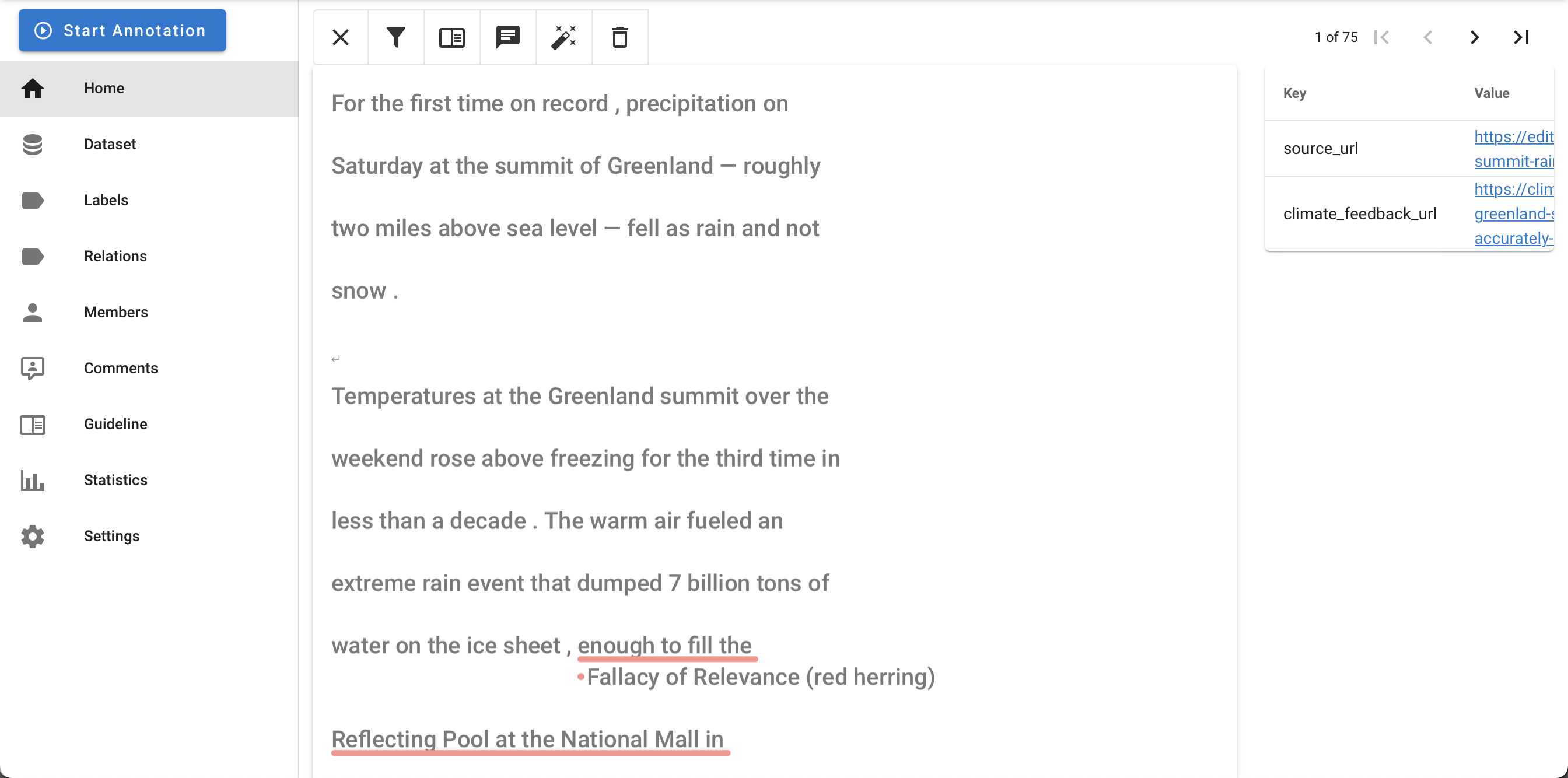}
    \caption{Annotation interface for the \logicClimate{} challenge set.}
    \label{fig:annot}
\end{figure}

\begin{figure}[h]
    \centering
    \includegraphics[width=\columnwidth]{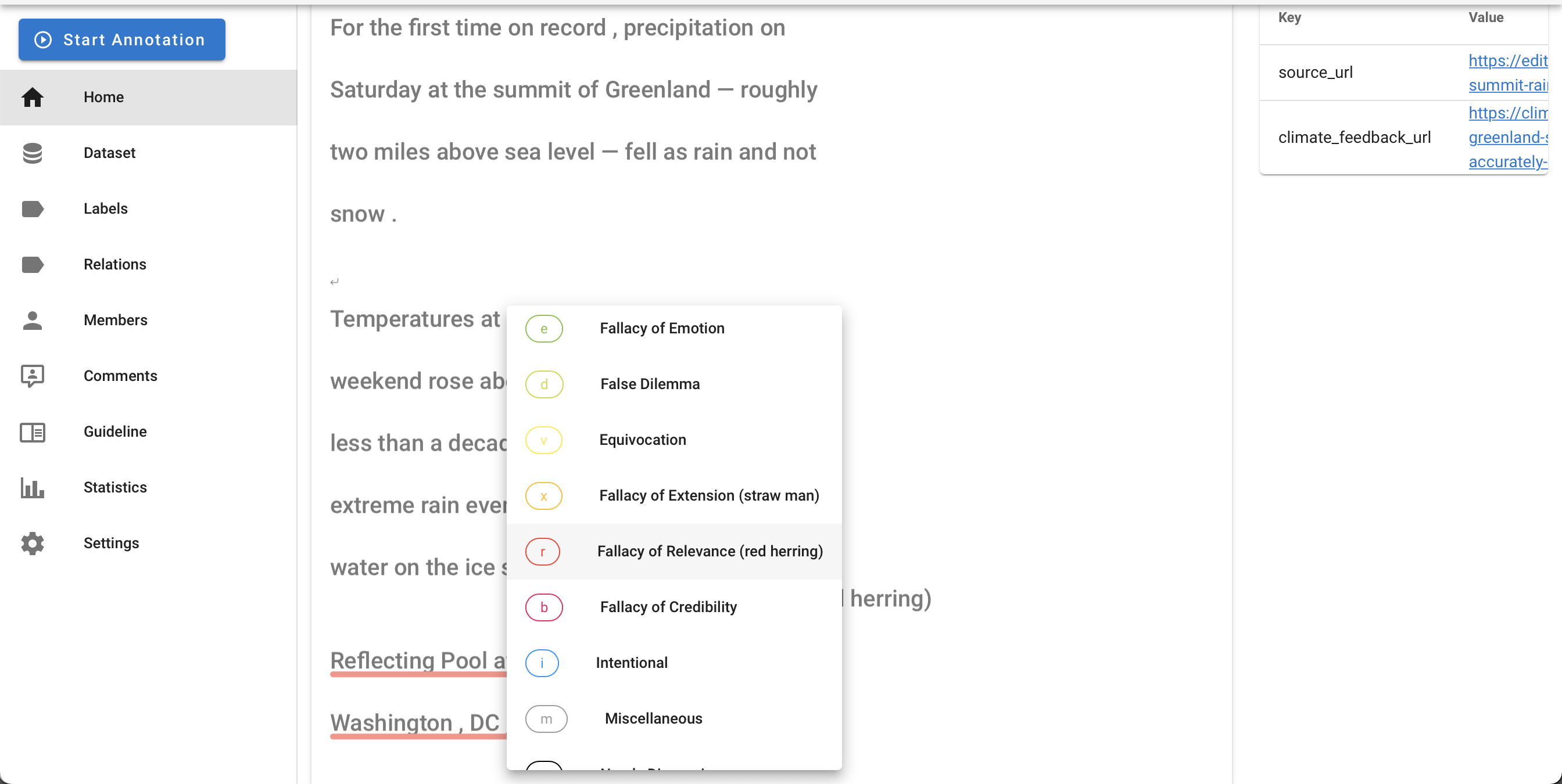}
    \caption{Choices of logical fallacy types in the annotation interface of the \logicClimate{} challenge set.}
    \label{fig:annot_labels}
\end{figure}

- Full text of instructions given to participants: We first show to the participants the description and examples of the 13 logical fallacy types as in \cref{appd:fallacy_type_details},
% \footnote{\url{https://docs.google.com/document/d/1-v_9a90OniMV8rOZqr7n0y3cRfBnROXJJG500EXTAzc/edit}} 
and when they are actually annotating, the interface screenshots are in \cref{fig:annot,fig:annot_labels}.
% , including, e.g., screenshots, disclaimers of any risks to participants or annotators, etc.
\myparagraph{Data sheet:}

- Why was the dataset created: We created the dataset for the proposed logical fallacy classification task.

- Who funded the creation of the dataset: The \logicGeneral{} part was collected by the co-authors, and the \logicClimate{} part was collected using the funding of a professor at the university.

- What preprocessing/cleaning was done: We tokenized the text using the word tokenization function of NLTK.\footnote{\url{https://nltk.org/}}
% (e.g., discretization or bucketing, tokenization, part-of-speech tagging, SIFT feature extraction, removal of instances)

% - If it relates to people, were they told what the dataset would be used for and did they consent? If so, how? Were they provided with any mechanism to revoke their consent in the future or for certain uses?

- Will the dataset be updated; how often, by whom: No, the dataset will be fixed.

\myparagraph{Additional ethical concerns:}

- Whether the data that was collected/used contains any information that names or uniquely identifies individual people or offensive content: No, the dataset does not contain personal information.

- License or terms for use and/or distribution: The dataset is open-sourced with the MIT license, and the intended use is for academic research but not commercial purposes.

% - If your use of existing artifact(s) was consistent with their intended use, provided that it was specified? For the artifacts you create, do you specify intended use and whether that is compatible with the original access conditions (in particular, derivatives of data accessed for research purposes should not be used outside of research contexts)?

% - potential risks
\begin{table*}[t]
\centering\small
\begin{tabular}{ p{0.15\linewidth} p{0.4\linewidth} p{0.4\linewidth}}

\toprule
\textbf{Fallacy Name} & \textbf{Description} & \textbf{Logical Form} \\
\midrule
Faulty Generalization & An informal fallacy wherein a conclusion is drawn about all or many instances of a phenomenon on the basis of one or a few instances of that phenomenon. is an example of jumping to conclusions. & [MSK1] has attribute [MSK2]. [MSK1] is a subset of [MSK3].
Therefore, all [MSK3] has attribute [MSK2]. (\href{https://en.wikipedia.org/wiki/Faulty_generalization}{Reference}) \\ \hline
False Causality & A statement that jumps to a conclusion implying a causal relationship without supporting evidence &[MSK1] occurred, then [MSK2] occurred. Therefore, [MSK1] caused [MSK2]. (\href{https://en.wikipedia.org/wiki/Post_hoc_ergo_propter_hoc}{Reference}) \\ \hline
Circular Claim & A fallacy where the end of an argument comes back to the beginning without having proven itself. & [MSK1] is true because of [MSK2]. [MSK2] is true because of [MSK1]. (\href{https://en.wikipedia.org/wiki/Circular_reasoning}{Reference}) \\ \hline
Ad Populum & A fallacious argument which is based on affirming that something is real or better because the majority thinks so. & A lot of people believe [MSK1]. Therefore, [MSK1] must be true. (\href{https://www.logicallyfallacious.com/logicalfallacies/Appeal-to-Common-Belief}{Reference}) \\ \hline
Ad Hominem & An irrelevant attack towards the person or some aspect of the person who is making the argument, instead of addressing the argument or position directly. & [MSK1] is claiming [MSK2]. [MSK1] is a moron. Therefore, [MSK2] is not true. (\href{https://www.logicallyfallacious.com/logicalfallacies/Ad-Hominem-Abusive}{Reference}) \\ \hline
Deductive Fallacy & An error in the logical structure of an argument. & If [MSK1] is true, then [MSK2] is true. [MSK2] is true. Therefore, [MSK1] is true. (\href{https://en.wikipedia.org/wiki/Formal_fallacy}{Reference}) \\ \hline
Appeal to Emotion & Manipulation of the recipient's emotions in order to win an argument. & [MSK1] is made without evidence. In place of evidence, emotion is used to convince the interlocutor that [MSK1] is true. (\href{https://www.logicallyfallacious.com/logicalfallacies/Appeal-to-Emotion}{Reference}) \\ \hline
False Dilemma & A claim presenting only two options or sides when there are many options or sides. & Either [MSK1] or [MSK2] is true. (\href{https://www.logicallyfallacious.com/logicalfallacies/False-Dilemma}{Reference}) \\ \hline
Equivocation & An argument which uses a key term or phrase in an ambiguous way, with one meaning in one portion of the argument and then another meaning in another portion of the argument. & [MSK1] is used to mean [MSK2] in the premise. [MSK1] is used to mean [MSK3] in the conclusion. (\href{https://www.logicallyfallacious.com/logicalfallacies/Equivocation}{Reference}) \\ \hline
Fallacy of Extension & An arugment that attacks an exaggerated or caricatured version of your opponent's position. & [MSK1] makes claim [MSK2]. [MSK3] restates [MSK2] (in a distorted way). [MSK3] attacks the distorted version of [MSK2]. Therefore, [MSK2] is false. (\href{https://www.logicallyfallacious.com/logicalfallacies/Strawman-Fallacy}{Reference}) \\ \hline
Fallacy of Relevance & Also known as red herring, this fallacy occurs when the speaker attempts to divert attention from the primary argument by offering a point that does not suffice as counterpoint/supporting evidence (even if it is true). & It is claimed that [MSK1] implies [MSK2], whereas [MSK1] is unrelated to [MSK2] (\href{https://www.txstate.edu/philosophy/resources/fallacy-definitions/Red-Herring.html}{Reference}) \\ \hline
Fallacy of Credibility & An appeal is made to some form of ethics, authority, or credibility. &
[MSK1] claims that [MSK2].
[MSK1] are experts in the field concerning [MSK2].
Therefore, [MSK2] should be believed.
% [MSK1] makes [MSK2]. [MSK3] claims [MSK1] is not a credible source regarding [MSK2]. Therefore, [MSK2] is false
(\href{https://en.wikipedia.org/wiki/Argument_from_authority}{Reference}) \\ \hline
Intentional Fallacy & 
A custom category for when an argument has some element that shows intent of a speaker to win an argument without actual supporting evidence.
% Some intentional (sometimes subconscious) action/choice to incorrectly support an argument.
& [MSK1] knows [MSK2] is incorrect. [MSK1] still claim that [MSK2] is correct using an incorrect argument. \\
\bottomrule
\end{tabular}
\caption{Types of logical fallacies along with their descriptions and logical forms.}
\label{tab:each_fallacy_desc}
\end{table*}

\subsection{Data Filtering Details of \logicGeneral{}}\label{appd:logic_general_filtering}
The data automatically crawled from quiz websites contain lots of noises, so we conducted multiple filtering steps. The raw crawling by keyword matching such as ``logic'' and ``fallacy'' gives us 52K raw, unclean data samples, from which we filtered to 1.7K clean samples.

As not all of the automatically retrieved quizzes are in the form of ``Identify the logical fallacy in this example: [...]'', we remove all instances where the quiz question asks about irrelevant things such as the definition of a logical fallacies, or quiz questions with the keyword ``logic'' but in the context of other subjects such as logic circuits for electrical engineering, or pure math logic questions. This is done by writing several matching patterns. After several processing steps such as deleting duplicates, we end up with 7,389 quiz questions. Moreover, as there is some noise that cannot be easily filtered by pattern matching, we also manually go through the entire dataset to only keep sentences that contain examples of logical fallacies, but not other types of quizzes. 
% len(raw_data): 90982
% len(raw_data): 90982
% len(raw_data): 90982
% len(non_empty_data): 29794
% len(clean_data): 7821
% [Info] Writing 7821 lines into data/sentence_fallacy_data.csv
% [Info] Writing 81882 lines into data/sentence_fallacy_deleted.csv
% [Info] Read 7821 lines from data/sentence_fallacy_data.csv
% len(clean_data): 7821
% len(clean_data): 7389
% [Info] Writing 1669 lines into data/sentence_fallacy_data_normed.csv
% len(clean_data): 1669
% finished normalization
% [Info] Read 1669 lines from data/sentence_fallacy_data_normed.csv
% [Info] Writing 1669 lines into data/sentence_logic_quiz.csv
% [Info] Read 1669 lines from data/sentence_logic_quiz.csv

The entire cleaning process resulted in 1.7K high-quality logically fallacious claims in our dataset.
% We open-source the dataset at \url{https://github.com/causalNLP/logical-fallacy}. 
As a reference, for each fallacy example we also release the URL of the source website where we extract this example from.

\subsection{Logical Fallacy Types} \label{appd:logic_types}
As different sources use different names for logical fallacies, we composed a set of 13 logical fallacy categories by conforming to set of logical fallacies given by Wikipedia,\footnote{\url{https://en.wikipedia.org/wiki/List_of_fallacies}} and considering the most common types in the dataset.
Therefore, we merged different surface forms of the same logical fallacy by listing out the different names of the same logical fallacy introduced on Wikipedia and also provided by educational websites. This leads to a reduction in logical fallacy types. For example, ``hasty induction'' and ``jumping to conclusions'' are merged under 
the category of ``hasty generalization.'' We further improve the eventual list by handcrafted rules, and delete data samples that cannot be matched to any of the logical fallacy types in our list. For a small number of remaining logical fallacy names which we cannot resolve automatically, we ask human annotators to align the names.

We introduce the detailed description for each of the 13 logical fallacy types and their logical forms in \cref{tab:each_fallacy_desc}. Most the descriptions and logical forms are collected from online websites introducing these logical fallacies. We provide the link of the source websites as references, and in some cases, we paraphrase the logical form to make it closer to natural text.
% \footnote{\url{https://bit.ly/3JECfHu}} 
In addition, we also provide the introduction of the 13 types that we compiled for the annotators in \cref{appd:fallacy_type_details}.
% ,\footnote{ \url{https://docs.google.com/document/d/1-v_9a90OniMV8rOZqr7n0y3cRfBnROXJJG500EXTAzc/}} and the explanations of how we created the 13 logical fallacy types.\footnote{\url{https://bit.ly/37eDDDy}}

\subsection{Data Annotation Details of \logicClimate{}}\label{appd:collection_logicclimate}

To ensure good annotation quality, we ask the annotators to pass a test batch of 65 samples after reading the definitions and examples of the 13 logical fallacies in the \logicGeneral{} dataset carefully. The test batch consists of 5 randomly selected samples for each of the 13 logical fallacies, and the annotators achieve above 85\% accuracy. We explained the examples where they did incorrectly and resolved their questions before they started annotating the \logicClimate{} test set.

Since each sample is annotated by two different annotators, we finalize the ground-truth labels in the following way: The two annotators merge all their annotations, and for places with divergent opinions, they cross-check with the experts' written reviews on the Climate Feedback website for each article. Specifically, each article is commented on by multiple expert reviewers such as professors, senior scientists and other researchers who explain what is fallacious with the article. If the labels can still not be unified after checking the expert reviews, since the annotators are trained to master the tasks very well (with 5+ hours of training, testing and discussions before the annotation), we let the two annotators have a discussion to decide the final label. 

\subsection{\logicClimate{} Examples}\label{appd:data_climate_examples}
We also show examples of \logicClimate{} in \cref{tab:data_climate}.    
\begin{table*}[t]
\centering \scriptsize
\begin{tabular}{p{1.5cm}p{13.5cm}}
\toprule
\textbf{Logical Fallacy} 
& \textbf{Examples}
\\ \hline
\textbf{Faulty Generalization}
& ``\textbf{For decades} horticulturalists have pumped carbon dioxide into glasshouses to increase yields. The fossil \textbf{record shows} that a thriving and diversification of plant and animal life \textbf{occurs every time} the atmosphere had a very high carbon dioxide content. \textbf{In the past, warming has never been a threat} to life on Earth.''
\\
\hline

\textbf{Ad Hominem}
& ``While CO2 levels were continuing to rise, temperatures weren’t. Hence the need for a \textbf{fallback position} — an environmental theory which would justify the massively expensive and disruptive ongoing decarbonisation programme so \textbf{assiduously championed} by politicians, scientists, green campaigners and anyone \textbf{making money out of the renewables business}. Ocean acidification fitted the bill perfectly.''
\\
\hline

\textbf{Ad Populum}
& ``According to a recent National Economic Research Associates Economic Consulting study, the Paris Agreement \textbf{could obliterate \$3 trillion} of GDP, \textbf{6.5 million industrial sector jobs} and \textbf{\$7,000 in per capita household income} from the American economy by 2040. Meeting the 2025 emissions reduction target alone \textbf{could subtract \$250 billion from our GDP} and \textbf{eliminate 2.7 million jobs}. The cement, iron and steel, and petroleum refining industries could see their \textbf{production cut by 21\% 19\%, and 11\%} respectively.''
\\
\hline

\textbf{False Causality}
& ``But like most claims regarding global warming, the real effect is small, \textbf{probably} temporary, and \textbf{most likely due} to natural weather patterns. Any changes in hurricanes over 70 years, even if real, \textbf{can easily be part of} natural cycles — or incomplete data. Coastal lake sediments along the Gulf of Mexico shoreline from 1,000 to 2,000 years ago \textbf{suggest} more frequent and intense hurricanes than occur today.''
\\
\hline

\textbf{Circular Claim}
& ``Even \textbf{if} enough accurate surface temperature measurements existed to ensure reasonable planetary coverage (\textbf{it doesn’t}) and to calculate some sort of global temperature statistic, interpreting \textbf{its significance would be challenging}. What averaging rule would you use to handle the data from thousands of temperature-sensing stations?''
\\
\hline

\textbf{Appeal to Emotion}
& ``There are now, trapped in Arctic ice, \textbf{diseases} that have not circulated in the air for millions of years — in some cases, since before humans were around to encounter them. Which means our immune systems would have \textbf{no idea how to fight back} when those prehistoric \textbf{plagues emerge} from the ice.''
\\
\hline

\textbf{Fallacy of Relevance}
& ``But \textbf{there are also reasons to believe} that environmental alarmism will, if not come to an end, have diminishing cultural power. The coronavirus pandemic \textbf{is an actual} crisis \textbf{that puts} the climate “crisis” \textbf{into perspective}. Even if you think we have overreacted, Covid-19 has killed nearly 500,000 people and shattered economies around the globe.''
\\
\hline

\textbf{Deductive Fallacy}
& ``Indeed, Queensland’s 2014 heat wave paled in comparison to the 1972 heat wave that occurred 42 years of global warming ago. \textbf{If} global warming \textbf{caused} the 2014 Queensland heat wave, \textbf{why wasn’t} it as severe as the 1972 Queensland heat wave? \textbf{Blaming every} single summer heat wave or extreme weather \textbf{event} on global warming is a stale and discredited tactic in the alarmist playbook.''
\\
\hline

\textbf{Intentional Fallacy}
& ``The bottom line is \textbf{there's no solid connection} between climate change and many major indicators of extreme weather that politicians keep talking about, such as hurricanes, tornadoes, droughts, rainfall and floods, despite Trudeau's claims to the contrary. The continual \textbf{claim of such links is misinformation} employed for political and rhetorical purposes.''
\\
\hline

\textbf{Fallacy of Extension}
& ``\textbf{For global warming alarmists}, however, a greener biosphere \textbf{is terrible} news and something \textbf{to be opposed}. This, in a nutshell, \textbf{defines the opposing sides} in the global warming debate. \textbf{Global warming alarmists claim} a greener biosphere with richer and more abundant plant life is horrible and justifies massive, economy-destroying energy restrictions. Global warming realists understand that a greener biosphere with richer and more abundant plant life \textbf{is not a horrible thing} simply because humans may have had some role in creating it.''
\\
\hline

\textbf{False Dilemma}
& ``America is poised to become a net energy exporter over the next decade. We \textbf{should not abandon that} progress \textbf{at the cost of} weakening our energy renaissance and crippling economic growth.''
\\
\hline

\textbf{Fallacy of Credibility}
& ``\textbf{I note particularly} that sea-level rise is not affected by the warming; it continues at the same rate, 1.8 millimeters a year, \textbf{according to} a 1990 review by Andrew S. Trupin and John Wahr. \textbf{I therefore conclude}—contrary to the general wisdom—that the temperature of sea water has no direct effect on sea-level rise. That means neither does the atmospheric content of carbon dioxide.''
\\
\hline

\textbf{Equivocation}
& ``Also, the alarmist assertion that polar ice sheets are melting is simply false. Although alarmists frequently point to a modest recent \textbf{shrinkage in the Arctic} ice sheet, that decline has been completely offset by ice sheet \textbf{expansion in the Antarctic}. \textbf{Cumulatively, polar ice sheets have not declined} at all since NASA satellite instruments began precisely measuring them 35 years ago.''
\\

\bottomrule
\end{tabular}
\caption{Examples of the \logicClimate{} data.}\label{tab:data_climate}
\end{table*}

\section{Implementation Details}

\subsection{Details of Zero-Shot Baselines}\label{appd:zero}

For the zero-shot classification models, we used pretrained NLI models \citep{yin-etal-2019-benchmarking} that are default choices of zero-shot classifiers in the transformers library \cite{wolf-etal-2020-transformers}: BART-MNLI and RoBERTa-MNLI.
For the implementation, we follow the standard pipeline introduced by huggingface.\footnote{\url{https://bit.ly/3E92Mvq}}

For the TARS model \citep{halder-etal-2020-task}, we follow the official documentation.\footnote{\url{https://github.com/flairNLP/flair/blob/master/resources/docs/TUTORIAL_10_TRAINING_ZERO_SHOT_MODEL.md}} All the above zero-shot classifiers show reasonable performance on existing datasets.\footnote{
% \url{https://nlp.town/blog/zero-shot-classification/}}  (by huggingface and flair), their accuracies evaluated on other datasets are reported at 
\url{https://github.com/nlptown/nlp-notebooks/blob/master/Zero-Shot\%20Text\%20Classification.ipynb}}

For GPT-3, we follow the official guide,\footnote{\url{https://beta.openai.com/docs/api-reference/classifications}} and use the following prompt (without additional efforts on prompt tuning because we do not assume any training samples for the zero-shot classification model):
% For the baselines, note that we only have the random choice baseline, because there should not be a baseline just always voting for the majority class, because we assume unseen training set, and there is no way to obtain the majority class without seeing the training set.
``Please classify a piece of text into the following categories of logical fallacies: [a list of all logical fallacy types].

Text: [Input text]

Label:''

We use the default search engine ``davinci,'' and the model ``curie.'' For reproducibility, we set the temperature to 0 for GPT-3 and all our zero-shot classification codes use a random seed of 1.

\subsection{Details of Finetuned Models}\label{appd:finetune_exp}
All models are finetuned using the NLI task, as motivated in \cref{sec:backbone}. We used a learning rate of $2^{-5}$, and the AdamW optimizer. We do not turn the learning rate and optimizers since the loss converges smoothly in all cases.
We tune the weights of the positive class, and set it to be 12, and the negative classes to be 1. The models used in our experiments have between 11M and 140M parameters. 
We train all models using NVIDIA TITAN RTX machines for less than two GPU Hours. 
For reproducibility, we fix the random seed to zero, and report the statistics of a single run.

% At inference time, the NLI process is repeated for each class label. We map the ``entailment'' class of NLI to ``fallacy'', and ``contradiction'' and ``neutral'' to ``non-fallacy.'' During inference time, we make a prediction over the 3 categories for each class by choosing the argmax of the model's predicted probabilities.

Due to the different dataset nature, our \logicGeneral{} is a single-label multi-class classification and the \logicClimate{} is a multi-label multi-class classification.

% 1st round of LM experiments:
% We used CUDA 10.1, and single-GPU, for about 30 min. We pretrained the models on MNLI for 3 hours for distilbert, 5 hours for albert, 12 hours for ELECTRA and most models, 14 hours big bird, 21 for deberta. Operating system is CentOS Linux 7 (Core).

% Requirements by ARR:
% GPU/CPU models; amount of memory; operating system; names and versions of relevant software libraries and frameworks. (hyper-)parameters used for each model/algorithm in the paper’s experiments. number and range of values tried per (hyper-)parameter during development of the paper, along with the criterion used for selecting the final parameter setting.
% the number of algorithm runs used to compute each reported result.

% measures of variation, confidence, or other distributional information.

\subsection{Details of Our Structure-Aware Model}\label{appd:implementation_our_model}
For the structure-aware classifier, we set the threshold of cosine similarity between two text spans to be 0.7, which is tuned using a manual grid search based on performance on the dev set.
In the training, we keep the training samples of the original text, and add additional samples using the masked text; in the inference stage, we choose the input format that performs the better on the development set, which is the masked text format. 

\section{Additional Experiments}
\subsection{Class-Specific Performance on \logicClimate{}}\label{appd:finetune_exp_lc}

\begin{table}[h]
\centering\small
\setlength\tabcolsep{5pt}
\setlength\extrarowheight{2pt}
\begin{tabular}{lcccc}
\toprule
{}  &  $F_1$ &     P &      R & Freq.
% Test \% 
\\
\midrule
Intentional & 24.58 & 100.00 & 39.46 & 25.58 \\
Appeal to Emotion & 23.40 & 84.62 & 36.67 & 11.37 \\
Faulty Generalization & 16.56 & 96.43 & 28.27 & 10.18 \\
Fallacy of Credibility & 25.00 & 45.00 & 32.14 & 9.90\\
Ad Hominem & 41.67 & 66.67 & 51.28 & 7.84\\
Fallacy of Relevance & 12.73 & 31.82 & 18.18 & 7.80\\
Deductive Fallacy & 9.32 & 64.71 & 16.30 & 6.50\\
False Causality & 15.15 & 31.25 & 20.41 & 5.11\\
Fallacy of Extension & 0.00 & 0.00 & 0.00 & 4.91\\
Ad Populum & 0.00 & 0.00 & 0.00 & 4.55\\
False Dilemma & 16.67 & 16.67 & 16.67 & 3.80\\
Equivocation & 5.00 & 20.00 & 8.00 & 1.94\\
Circular Claim & 0.00 & 0.00 & 0.00 & 0.51\\

\hline
Overall & \textbf{29.37} 
& {17.66} & 67.22 & 8.37 \\
\bottomrule
\end{tabular}
\caption{Class-specific performance achieved by \ourmodel{} on \logicClimate{}. For each class, we report the $F_1$ score, precision (P), recall (R), and the frequency (Freq.) of the class in the \logicClimate{} dataset. Note that the Freq. column is copied from \cref{tab:label_freq}.}
\label{tab:lc_classw}
\end{table}
For \logicClimate{}, we provide class-specific performance for the best performing model in \cref{tab:lc_classw}. There is lots of space for future work to improve the performance on this dataset. For error analysis with the current best performing model, we identify the following aspects that make \logicClimate{} more challenging than \logicGeneral{}.
We measure the complexity and diversity of the dataset by measuring the BLEU score difference with the logical forms, and find that \logicClimate{} (0.18) has a lower similarity than \logicGeneral{} (0.24). The relatively higher complexity and diversity might explain why the best model only achieves 29.37\% on \logicClimate{}. We also find that as \logicGeneral{} has examples that are designed such that students can classify them, \logicClimate{} has fallacies created by top-level journalists created with the intention that even educated readers will not be able to detect them. The small size of the dataset might also be a factor, as we find that the best performing model achieves a similar performance (34.52\%) on \logicGeneral{} when trained on the same amount of data. We also find that the model struggles on classes with small amounts of data.

\subsection{Finetuning on \logicClimate{}}
\label{appd:logic_climate_finetune}

To obtain the performance of \ourmodel{} vs. \electra{} after finetuning on the \logicClimate{} dataset, we split the \logicClimate{} dataset into train, dev, and test splits. Dataset statistics are shown in \cref{tab:climate_stats}.
\begin{table}[h]
    \centering
    %\small
      \resizebox{\columnwidth}{!}{%
    \begin{tabular}{lccccc}
    \toprule
     & \textbf{\# Samples} & \textbf{\# Sents} & \# \textbf{Tokens} & \textbf{Vocab}  \\ \hline
\textbf{Total Data} & 1,079 & 1,463 & 38,828 & 5,809 \\
\quad Train & 680 & 891 & 24,814 & 4,402 \\
\quad Dev & 219 & 331 & 8,419 & 2,229 \\
\quad Test & 180 & 241 & 5,595 & 1,707 \\
    \bottomrule
    \end{tabular}
    }
    \caption{Statistics of the \logicGeneral{} dataset.}
    \label{tab:climate_stats}
\end{table}

\section{Details of All Fallacy Types}\label{appd:fallacy_type_details}

We list the details of all fallacy types below. We also use this list to guide annotators to identify logical fallacies.
\paragraph{Faulty Generalization}

\begin{itemize}[nolistsep]
\item
  \textbf{Definition:} This fallacy occurs when an argument applies a
  belief to a large population without having a large enough sample to
  do so.
\item
  \textbf{Example:} A New York driver cuts you off in traffic. You then
  decide that all New Yorkers are terrible drivers.
\item
  \textbf{Synonyms or Subtypes:} Slippery Slope, Hasty Generalization,
  Accident, Fallacy of Division, Error of Division, Error of
  Composition, Property in the Whole, Property in the Parts, Causal
  Oversimplification, Part to Whole, Association Fallacy, Guilt by
  Association, Composition Fallacy, Ecological Fallacy, Conjunction
  Fallacy, False Analogy, Inconsistent Comparison, Package Deal,
  Overwhelming Exception, False Equivalence, All Things Are Equal,
  McNamara Fallacy.
\end{itemize}

\paragraph{False Causality}

\begin{itemize}[nolistsep]
\item
  \textbf{Definition:} This fallacy occurs when an argument assumes that
  since two events are correlated, they must also have a cause and
  effect relationship.
\item
  \textbf{Example:} We observed an increase in ice cream sales at the
  same time as air conditioner sales increased. Therefore, we can
  conclude that selling more ice cream causes more air conditioners to
  be sold.
\item
  \textbf{Synonyms or Subtypes:} Post hoc ergo propter hoc, Cum hoc ergo
  propter hoc, Regression Fallacy, Consecutive Relation, Magical
  Thinking, Gambler's Fallacy (rarely called temporal flaw/temporal
  fallacy), Ludic Fallacy.
\end{itemize}

\paragraph{Circular Claim}

\begin{itemize}[nolistsep]
\item
  \textbf{Definition:} This fallacy occurs when an argument uses the
  claim it is trying to prove as proof that the claim is true.
\item
  \textbf{Example:} You must obey the law, because it is illegal to
  break the law.
\item
  \textbf{Synonyms or Subtypes:} Circular Reasoning, Homunculus Fallacy.
\end{itemize}

\paragraph{Ad Populum}

\begin{itemize}[nolistsep]
\item
  \textbf{Definition:} This fallacy occurs when an argument is based on
  affirming that something is true because a statistical majority
  believes so.
\item
  \textbf{Example:} Most people believe that there is a God, therefore
  it must be true.
\item
  \textbf{Synonyms or Subtypes:} Appeal to the Public, Ad Numerum,
  Appeal to the Numbers, Bandwagon Fallacy.
\end{itemize}

\paragraph{Ad Hominem}

\begin{itemize}[nolistsep]
\item
  \textbf{Definition:} This fallacy occurs when a speaker trying to
  argue the opposing view on a topic makes claims against the other
  speaker instead of the position they are maintaining.
\item
  \textbf{Example:} Person A makes a claim. Person B says that Person
  A's claim is false because Person A is not a hard worker.
\item
  \textbf{Synonyms or Subtypes:} Genetic Fallacy, Tu quoque (you too),
  Bulverism, Poisoning the Well, Appeal to Hypocrisy, Traitorous Critic.
\end{itemize}

\paragraph{Deductive Fallacy}

\begin{itemize}[nolistsep]
\item
  \textbf{Definition:} This fallacy occurs when there is a logical flaw
  in the reasoning behind the argument, such as a propositional logic
  flaw.
\item
  \textbf{Example:}

  \begin{itemize}[nolistsep]
  \item
    Affirming the consequent: If A is true then B is true. B is true.
    Therefore, A is true.
  \item
    Denying the antecedent: If A is true then B is true. A is false.
    Therefore, B is false.
  \item
    Affirming a disjunct: A or B is true. B is true. Therefore, A is not
    true.
  \end{itemize}
\item
  \textbf{Synonyms or Subtypes:} False Analogy, Affirming the
  Consequent, Non-sequitur, Four Terms Fallacy, Affirming the Disjunct,
  Argument From Fallacy (correct identification of fallacy, but
  incorrect conclusion), Appeal to Probability, Undistributed Middle,
  Moral Equivalence, Self contradiction, Internal Contradiction,
  Masked-man Fallacy, Four Terms, Illicit Major, Illicit Minor, Denying
  the Antecedent, Existential Fallacy, Kettle Logic,
  \hspace{0pt}\hspace{0pt}Affirmative Conclusion from a Negative
  Premise, Negative Conclusion from a Negative Premise, Exclusive
  Premises.
\end{itemize}

\paragraph{Appeal to Emotion}

\begin{itemize}[nolistsep]
\item
  \textbf{Definition:} This fallacy is when emotion is used to support
  an argument, such as pity, fear, anger, etc.
\item
  \textbf{Example:} You should marry me. I know we're not compatible,
  but you're my last chance.
\item
  \textbf{Synonyms or Subtypes:} Appeal to Pity, Appeal to Fear, Ad
  baculum (appeal to force), Appeal to Ridicule, Appeal to Gallery,
  Wishful Thinking, Appeal to Consequences, Appeal to Spite, Appeal to
  Force, Appeal to Flattery.
\end{itemize}

\paragraph{False Dilemma}

\begin{itemize}[nolistsep]
\item
  \textbf{Definition:} This fallacy is when incorrect limitations are
  made on the possible options in a scenario when there could be other
  options.
\item
  \textbf{Example:} You're either for the war or against the troops.
\item
  \textbf{Synonyms or Subtypes:} Either/Or thinking, Black-or-White
  Fallacy, False Dichotomy, Nirvana Fallacy, Perfect Solution.
\end{itemize}

\paragraph{Equivocation}

\begin{itemize}[nolistsep]
\item
  \textbf{Definition 1:} This fallacy can occur in two ways: the first
  is when ambiguous or evasive language is used to avoid committing
  oneself to a position.
\item
  \textbf{Example:}
\end{itemize}

Speaker 1: Did you torture the prisoner?

Speaker 2: No, we just held him under water for a while, and then did a
mock hanging.

\begin{itemize}[nolistsep]
\item
  \textbf{Definition 2:} The second way equivocation occurs is when the
  same is word is used in an argument but with different meanings:
\item
  \textbf{Example 2:}
\end{itemize}

Speaker 1: We are using thousands of people to go door to door and help
spread the word about social injustice and the need for change.

Speaker 2: I can't be a part of this because I was taught that using
people is wrong.

\begin{itemize}[nolistsep]
\item
  \textbf{Definition 3:} An equivocation seeks to draw comparisons
  between different, often unrelated things.
\item
  \textbf{Synonyms or Subtypes:} Uncertain use of term or concept,
  Reification, Continuum fallacy, False attribution, Moral equivalence,
  Etymological Fallacy.
\end{itemize}

\paragraph{Fallacy of Extension}

\begin{itemize}[nolistsep]
\item
  \textbf{Definition:} Also known as straw man, this is when an argument
  appears to be refuted by being replaced with an argument with a
  similar but weaker argument.
\item
  \textbf{Example:}
\end{itemize}

Speaker 1: I think we should have single payer, universal, healthcare.

Speaker 2: Communist countries tried that. We don't want America to be a
communist country so we shouldn't have single payer healthcare.

\begin{itemize}[nolistsep]
\item
  \textbf{Synonyms or Subtypes:} Straw man, Suppressed Correlative.
\end{itemize}

\paragraph{Fallacy of Relevance}

\begin{itemize}[nolistsep]
\item
  \textbf{Definition:} Also known as red herring, this fallacy occurs
  when the speaker attempts to divert attention from the primary
  argument by offering a point that does not suffice as
  counterpoint/supporting evidence (even if it is true).
\item
  \textbf{Example:} We should move our office to California to expand
  our potential customers. And the weather is warmer there, which is all
  the more reason to move there.
\item
  \textbf{Synonyms or Subtypes:} Red herring, Two wrongs make a right,
  Argument to moderation, Moralistic fallacy, Moral equivalence, Logic
  chopping, Proof by assertion, Argument from silence, Irrelevant
  material, Relative privation.
\end{itemize}

\paragraph{Fallacy of Credibility}

\begin{itemize}[nolistsep]
\item
  \textbf{Definition:} This fallacy is when an appeal is made to some
  form of ethics, authority, or credibility.
\item
  \textbf{Example:} If mailing a hand-written letter was good enough in
  the past, then you don't need those pesky computers (appeal to
  tradition).
\item
  \textbf{Synonyms or Subtypes:} Appeal to authority, Appeal to to
  nature, Naturalistic fallacy, Appeal to tradition, Chronological
  snobbery (reverse of tradition), Appeal to novelty, Ipse dixit,
  Etymological fallacy, Appeal to poverty, Appeal to accomplishment.
\end{itemize}

\paragraph{Intentional Fallacy}

\begin{itemize}[nolistsep]
\item
  \textbf{Definition:} This is sort of a ``custom-made'' category for
  when an argument has some element that shows ``intent'' of a speaker
  to win an argument without actual supporting evidence.
\item
  \textbf{Example:} Can you meet to discuss this tomorrow, or are you
  too busy slacking off? (loaded question - the person who answers with
  yes/no is cornered into discussing or slacking off)
\item
  \textbf{Extra example:} A dating app matches Joe and Jane because they
  both love the same shows, music, and going to the beach. It did not
  take into account their 40 year age difference, or that Joe works
  overnight shifts and Jane works 9-5 (texas sharpshooter).
\item
  \textbf{Synonyms or Subtypes:} Texas sharpshooter, Cherry picking,
  Mcnamara fallacy, No true scotsman, Appeal to ignorance/argument from
  ignorance, Complex question, Moving the goalposts, Loaded question,
  Special pleading, Hiding information/half truth, Many questions,
  Incredulity, Divine Fallacy, Quoting out of context, Shifted burden of
  proof, Ambiguous words or phrases.
\end{itemize}

% \section{Classwise Distribution}

% \begin{table*}[t]
%     \centering\small
%     \begin{tabular}{lccccccccccccc}
% \toprule
% % \textbf{Logical Fallacy Type} & \textbf{Frequency in Data} \\ \hline
% title && ad hominem && ad populum && appeal to emotion && circular reasoning && equivocation && fallacy of credibility && fallacy of extension && fallacy of logic && fallacy of relevance && false causality && false dilemma && faulty generalization && intentional \\
% edu\_train && 12.17 && 8.55 && 7.03 && 7.25 && 2.11 && 5.79 && 5.73 && 6.54 && 6.17 && 9.41 && 5.95 && 17.25 && 6.06 \\
% edu\_dev && 12.00 && 14.67 && 4.67 && 6.00 && 1.67 && 2.67 && 4.67 && 5.67 && 8.00 && 8.00 && 6.33 && 20.33 && 5.33 \\
% edu\_test && 13.67 && 10.00 && 7.67 && 6.33 && 1.67 && 5.67 && 7.00 && 4.67 && 8.00 && 6.00 && 4.00 && 20.33 && 5.00 \\
% climate\_train && 8.07 && 5.96 && 10.88 && 0.70 && 2.22 && 10.99 && 3.98 && 6.55 && 7.25 && 4.44 && 4.09 && 10.06 && 24.80 \\
% climate\_dev && 6.50 && 2.53 && 10.83 && 0.00 && 1.44 && 6.86 && 9.39 && 5.78 && 7.22 && 6.86 && 3.61 && 8.30 && 30.69 \\
% climate\_test && 7.76 && 2.28 && 12.79 && 0.46 && 1.83 && 9.13 && 2.28 && 7.76 && 11.42 && 7.31 && 3.20 && 13.24 && 20.55 \\

% \bottomrule
%     \end{tabular}
%     \caption{Logical fallacy types and their frequencies in the \logicClimate{} dataset.}
%     \label{tab:label_freq}
% \end{table}

\end{document}